# Efficient Classification of Multi-Labelled Text Streams by Clashing

Ricardo Ñanculef[a,b], Ilias Flaounas[b], Nello Cristianini[b]

[a]*Department of Informatics, Universidad Técnica Federico Santa María, Chile.*
[b]*Intelligent Systems Laboratory, University of Bristol, UK.*

**Abstract**

We present a method for the classification of multi-labelled text documents explicitly designed for data stream applications that require to process a virtually infinite sequence of data using constant memory and constant processing time.

Our method is composed of an online procedure used to efficiently map text into a low-dimensional feature space and a partition of this space into a set of regions for which the system extracts and keeps statistics used to predict multi-label text annotations. Documents are fed into the system as a sequence of words, mapped to a region of the partition, and annotated using the statistics computed from the labelled instances colliding in the same region. This approach is referred to as *clashing*.

We illustrate the method in real-world text data, comparing the results with those obtained using other text classifiers. In addition, we provide an analysis about the effect of the representation space dimensionality on the predictive performance of the system. Our results show that the online embedding indeed approximates the geometry of the full corpus-wise TF and TF-IDF space. The model obtains competitive F measures with respect to the most accurate methods, using significantly fewer computational resources. In addition, the method achieves a higher macro-averaged F measure than methods with similar running time. Furthermore, the system is able to learn faster than the other methods from partially labelled streams.

## 1. INTRODUCTION

The efficient analysis of massive datasets is one of the main challenges in modern machine learning and data mining applications (Rajaraman and Ullman, 2012; Hand, 2013; Wu et al., 2014). Usually in these scenarios, data is being generated continuously, arriving to the system in the form of a fast and virtually infinite data stream (Aggarwal, 2007; Bifet, 2013). Examples include the stream of messages exchanged on a social

*Email addresses:* `jnancu@inf.utfsm.cl` (Ricardo Ñanculef), `ilias.flaounas@bristol.ac.uk` (Ilias Flaounas), `nello.cristianini@bristol.ac.uk` (Nello Cristianini)

network or the stream of daily stories generated by different news outlets. The challenge for a mining system designed to work under this setting is being ready to predict at any time, learning constantly from new observations, but using limited computational resources i.e. bounded memory and constant processing time.

In this paper, we present a simple and efficient method to classify large data streams of documents: one of the most common type of user generated data. Document classification is one of the most frequent and important problems in textual data analysis, with applications from information retrieval and spam filtering to content personalization and natural language processing. The task is that of learning a mechanism from data to automatically annotate documents with thematic categories or labels from a given set (Sebastiani, 2002; Aggarwal and Zhai, 2012). Since documents can be associated with multiple non-exclusive categories (e.g. *politics*, *economics* and *international affairs*) simultaneously, this task is one of the most common examples of multi-label classification (Tsoumakas et al., 2010).

The novelty and contribution of this work is in addressing explicitly and simultaneously text representation and multi-label classification in streaming environments with limited computational resources. As we explain below, this is a challenging setting because of the large number of possible features arising in textual domains and the traditional batch setting for dimensionality reduction and multi-label classifier design (Read et al., 2012).

*1.1. Context of this Research*

In this section we motivate the settings of this research, discussing current work in the field, and stressing the novel aspects of our approach.

*1.1.1. Text Representation*

Different methods to classify documents have been investigated in the last years (Aggarwal and Zhai, 2012; Sebastiani, 2002). A fundamental component of these systems is the way to represent text into an amenable form for machine learning algorithms. This representation is commonly obtained by selecting a set of indexing terms suitable to capture document content (the vocabulary) and a weighting scheme assigning values to the dimensions of the feature vector spanned them (Joachims, 2002; Zhang et al., 2011). The bag of words (BOW) indexing model is the most widely used text representation method in current research (Lan et al., 2009; Ren and Sohrab, 2013). In this model, each possible word in the set of known texts corresponds to a dimension of the feature space used to embedd documents. Along with BOW, the TF-IDF weighting scheme is usually applied to obtain the final representation of a document (Lan et al., 2009; Aggarwal and Zhai, 2012). TF-IDF is proportional to the number of times a particular word appeared in a document and inversely proportional to the number of documents containing the word. Despite its widespread acceptance among practitioners, this approach for text representation has some drawbacks that recently have started to be addressed by researchers in the field. First, documents are treated as collections of unordered words. A number of authors have thus investigated the use of longer indexing units, linguistically or statistically meaningful for content identification (Zhang et al., 2011). They include k-grams (Caropreso et al., 2001), frequent word sequences (Li et al., 2008) and frequent word sets (Zhang et al., 2010, 2011). Unfortunately, in text categorization problems, these methods have shown improvements somewhat disappointing (Li et al., 2011; Zhang et al., 2011). Second, TF-IDF does not exploit the co-occurrence of terms and categories in the weighting process. Therefore, methods capable to exploit information about the different distribution of terms among the documents of each class have been focus of increasing interest in the last years (Lan et al., 2009; Guan et al., 2009; Luo et al., 2011; Ren and Sohrab, 2013; Wang and Zhang, 2013) and is shaping up as an important direction of research. However, as regards the scalability of text classification systems, the most important drawback of BOW and TF-IDF is the high-dimensionality of the resulting representation space.

The dimensionality of TF-IDF matches the size of the vocabulary i.e. the number of distinct terms across the entire dataset. This "curse of dimensionality" in traditional text representation brings about huge memory requirements and huge computation since most classification models scale linearly or super linearly in the dimensionality of the feature set size. In data stream scenarios, the problem is still worse, because the word distribution (required to compute TF and IDF) is not known beforehand and both the vocabulary and the corpus is constantly growing. Recently, various feature selection techniques to reduce dimensionality have been studied and compared in text domains (Forman, 2003; Fragoudis et al., 2005; Yang et al., 2012; Spolaôr and Tsoumakas, 2013). Most of these methods correspond to filter approaches, that is, methods selecting features from general characteristics of the training data regardless of the learning algorithm (Spolaôr and Tsoumakas, 2013). Although wrapping methods, using the classifier to determine the quality of selected features, usually outperform filter methods, they tend to be prohibitively expensive on large-scale datasets (Yang et al., 2012; Wang et al., 2013b; Spolaôr and Tsoumakas, 2013) and thus more simple and efficient methods such as Information Gain, Chi-Squared and Bi-normal Separation (Forman, 2003) are preferred in practice. Unfortunately, despite the increasing importance of data stream scenarios, most studies on feature selection are restricted to the batch setting, that is, the selection task is conducted off-line and all the features and training instances are supposed to be known a priori. Thus, weighting schemes like TF-IDF and feature selection methods used to reduce dimensionality, need to be updated at a corpus level which demands lots of computation and requires the storage of a large amount of training data. A fully online feature selection method has been recently presented in (Wang et al., 2013b). However, it is focused on a binary single label classification with a perceptron classifier. Indeed, research on multi-label feature selection is still very scarce. Feature selection for text datasets often applies traditional filter methods focusing on a single label and then uses some aggregation strategy to obtain a decision. It is well known that this approach can neglect strongly predictive features for unfrequent labels in unbalanced categorization problems. Therefore, both multi label and online feature selection are still topics that need to be studied in the field.

In this paper we investigate a fully online *embedding* method for approximating TF-IDF using constant memory and time. In contrast to most dimensionality reduction approaches, we do not apply a reduction method on the original TF-IDF representation but directly on the sequence of words contained in a document. The method is built on the ideas of count min sketching (Cormode, 2012) and feature hashing (Shi et al., 2009b,a), methods introduced to estimate data stream distributions and high dimensional dot products respectively. Therefore, our method is related also to other data-oblivious embedding techniques like random projections (RP) (Achlioptas, 2003) and fast Johnson

Lindenstrauss transforms (FJL) (Ailon and Chazelle, 2010), for which there has been significant interest in the community. In contrast to these techniques, the tools we use do not rely on the application of cumbersome projection matrices to data but rely on simple hashing functions which can be directly applied to words. Random projections for text representation have been applied to text representation in (Lin and Gunopulos, 2003) and (DeBarr and Wechsler, 2012). Up to our knowledge, the computationally efficient versions of RP described in (Ailon and Chazelle, 2010) still have not been explicitly studied in text domains. Formerly, Baena-Garcia et al. (2011) has proposed using the count min sketch to allow the efficient computation of IDF for massive streams of documents, studying the similarity between the ranking of the exact TF-IDF values and that of the approximate values obtained from approximate IDF. However, this algorithm works with exact TF and authors do not assess (theoretically or empirically) the effects of this approximation on document classification tasks. Approximating TF using with fewer dimensions is important in text categorization problems, because classification algorithms usually rely on the computation of metrics which scale linearly in the number of dimensions of the representation space. In this work, we study the approximate computation of both TF and IDF studying both theoretically and empirically the quality of the approximation. Recently, a generic approach for mining massive data using sketches has been suggested in (Gupta et al., 2013). However it relies on low rank matrix multiplications and does not focus on text representation or text categorization in online environments.

*1.1.2. Multi-Label Text Classification*

Most classification methods studied in machine learning are devised to deal with single label assignments i.e. a data item belongs to one and only one class of the set of possible categories. Therefore, multi-label classification methods for problems arising in areas like text categorization, image annotation and protein function classification are of increasing research interest in the last years (Tsoumakas et al., 2010; Madjarov et al., 2012; Jiang et al., 2012; Yu et al., 2014; Montañes et al., 2014). Multi-label classification is usually approached using either a problem transformation approach, where the problem is decomposed into several classic classification tasks, or by directly designing a method to predict multiple classes at once (Madjarov et al., 2012; Yu et al., 2014). In the first category, the Binary Relevance model is the most widely used approach in practice (Tsoumakas et al., 2010). Other popular methods in text categorization include the Binary Pairwise model, Boostexter (based on the Adaboost algorithm) and BRkNN (based on kNN and the BR framework) (Madjarov et al., 2012). Recently, several papers have shown that classification accuracy can improve by taking into account the possible correlations among labels (Dembczyński et al., 2012a; Montañes et al., 2014). One of the first methods addressing this issue is the label powerset method that considers each distinct combination of labels that exist in the dataset as a different class value. Clearly, the computational complexity of this method is worst case and therefore several efforts have been directed towards making it more efficient (Read et al., 2008; Tsoumakas et al., 2011). The classifier chain method (CC) is another ingenious way to model label dependencies (Read et al., 2011) but still enjoying the advantages of BR. This method selects an order on the label set and trains a binary classifier for each label in this chain. Dependencies are included by extending the feature space of each classifier to include the label associations of all previous classifiers. Variants of this method have been recently proposed in (Dembczyński et al., 2012b), (Senge et al., 2013) and (Montañes et al., 2014).

Unfortunately, even if the problem of multi-label classification has seen considerable development in recent years, few authors have looked at this task in a data stream context (Read et al., 2012). Methods mentioned above suppose that both features and data are known beforehand. A straighTForward adaptation of them for online environments may thus require to periodically recompute a model to learn from new observations, an approach that require unbounded memory and time to store and process the full data stream. One possible solution at this level is exploiting problem transformation strategies and borrowing single label data stream classification techniques to obtain multi-labelled assignments (Read et al., 2011, 2012). However, this approach substantially limits the predictive performance and speed of the classification system due to the high imbalance the obtained subproblems which leads to overwhelm the class with more samples, the possible correlation between different labels and the large number of sub-models that needs to be updated in problem transformation schemes(Zhou et al., 2012). Up to our knowledge, the first method explicitly devised for multi-label classification of data streams has been recently proposed in (Read et al., 2012). This approach is based on the Very Fast Decision Tree (VFDT) introduced by Domingos and Hulten (2000) and extended several times in the literature (see e.g. Liang et al., 2012). A drawback of this method is that for numerical attributes VFDT's runtime is $\mathcal{O}(Zd)$ where $d$ is the number of attributes and $Z$ the set of possible splitting points. If $Z$ is large (e.g. $d$), complexity becomes prohibitive, but if $Z$ is small accuracy can significantly decrease.

In order to efficiently predict multi-label classifications, we propose to organize the feature space into a set of regions where documents with similar low dimensional representations collide by way of certain mapping process. Documents colliding in the same region are said *to clash*. Then, we implement a conditional naive bayesian approach. Labels are assumed to be independent given that the clashing region is known. This allows to model partial dependencies among labels and still keep the system efficient (Dembczyński et al., 2012b). A simple way to implement a tessellation of the feature space is by adopting a prototype based method. In particular, we focus on centroids as they have been successfully used in the text categorization literature and are generating a renewed interest in the last years due to their computational efficiency (Tan et al., 2011; Pang and Jiang, 2013; Wang et al., 2013a; Borodin et al., 2013). Tan (2008) reports significant improvements on naive Bayes and KNN methods using adaptive centroid classifiers in text categorization tasks. An online extension of this method has been formerly presented in (Tan et al., 2011) but applied to classic train/test problems where the method slightly outperforms SVMs. Recently, a similar technique was presented in (Borodin et al., 2013) for text classification in data stream environments. Unfortunately it focuses on single label classification and documents are represented using batch TF-IDF.

*1.2. Novelty and Contributions*

The main contribution of this paper is a simple and effective method for document classification where both components, text representation and classifier adaptation, are explicitly designed for large-scale multi-label data stream scenarios. The combined approach for representing and classifying text is referred to as *clashing*. This system efficiently extracts and keeps sufficient statistics for online text representation and multi-label classification, ensuring a truly bounded resources solution, where every operation is performed in constant time. As we explained in section 1.1.1, our setting and approach for text representation is novel and builds on recent ideas for mining massive data streams.

Novelty is here in the fact that both TF and IDF approximation are made entirely online and guarantee constant time and space. As we explain in section 1.1.2, our approach to obtain multi-label annotations is also novel and attempts to exploit and keep the efficiency of our online embedding method. It can be regarded as a multi-label centroid based classifier (CC) where multi-label annotations are obtained by adding (conditional) naive bayesian models at each region where documents *clash*. Novelty is here in the way to extend CC to multi-label data stream tasks. Below we provide some highlights about this work:

- We provide a document classification system where both text representation and multi-label annotation are online and guarantee constant processing time.

- We show (theoretically and empirically) that the online text representation system approximates the TF-IDF space without requiring corpus wise computations.

- We analyze the effects of the approximate representation on the classification system, taking exact TF-IDF as the baseline.

- We show that the proposed method is better or comparable, in terms of macro- and micro- averaged F measures, to a periodically recomputed SVM, but uses much fewer computational resources.

*1.3. Practical Implications of this Research*

The problem of classifying documents using finite resources is an important challenge in the domain of mining textual streams. Fast flowing streams of text are generated by online news, social media and endless other applications, and the need to automatically and adaptively sort them into sub-streams is a crucial one. The insistence on only making use of bounded resources is a consequence of the size of the streams: both time and memory need to be kept under control. We focused on this problem as part of our ongoing efforts in the analysis of web news, and the algorithm we have developed will be incorporated into our pipeline devoted to Computational Social Sciences (Flaounas et al., 2011).

*1.4. Organization of this paper*

The rest of this article is organized as follows. In section 2, we briefly introduce the problem of document classification and the data stream scenario. In this section we also present the hashing technique that will be used to obtain text representation in online domains. The clashing approach is described in section 3 with two variants to learn a partitioning of that space based on the available labels. In section 4 we present a theoretical analysis of the representation obtained by hashing compared to the geometry of the full word count space. Additional related work, not discussed in this introduction, is presented in section 5. In section 6, we present several experiments performed on the Reuters RCV1 corpus and additional scalability tests carried out in the New York Times dataset. The main conclusions and contributions of this paper are summarized in section 7. The appendix at the end of this manuscript provides the proofs of all the theoretical claims presented in section 5.

## 2. PROBLEM STATEMENT AND BACKGROUND

We denote by $\mathcal{D}$ the set of all possible documents and use $\mathbf{x}(d) \in \mathcal{X}$ to denote a vector representation $d \in \mathcal{D}$ obtained by using a text representation model $\mathbf{x} : \mathcal{D} \to \mathcal{X}$. We use $\mathbf{x}$ as a shorthand of $\mathbf{x}(d)$, and $\mathbf{x}_i$ as a shorthand of $\mathbf{x}(d_i)$, when this is clear from the context. A textual dataset or *corpus* is a collection of documents $D$ containing words $w_1, w_2, \ldots$ from a set $\mathcal{W}$.

### 2.1. Document Classification

Given a finite and non-empty set of $n_t$ labels $\mathcal{T} = \{\tau_1, \tau_2, \ldots, \tau_{n_t}\}$, the task of text classification is that of extracting from a dataset $D$ a decision mechanism $f : \mathcal{D} \to 2^{\mathcal{T}}$ that describes how documents ought be classified (Sebastiani, 2002), that is, how to annotate documents with labels which are relevant for them. Text classification can be seen as a kind of information retrieval task for each instance, where the labels play the role of the documents to retrieve (Quevedo et al., 2012). In contrast to *single-label* pattern classification tasks, text classification is a *multi-label* problem, that is, classes are not mutually exclusive: a document can be simultaneously assigned with several labels. In this paper, we assume that each document $d_i \in D$ has been annotated with class labels $T_i \subset \mathcal{T}$, that is, we adopt a supervised approach (Joachims, 2002), building $f$ both from $D$ and the known annotations $T_1, T_2, \ldots$.

Different methods for text classification have been explored in the last years, including linear models, nearest neighbor approaches, support vector machines (SVMs), neural networks, classifier ensembles and various generative methods (Sebastiani, 2002). Traditionally, previous work has focused on improving classification performance using classic train/test settings, which allows to employ sophisticated feature selection approaches to optimize text representation and data intensive training algorithms to obtain an accurate classifier. For a comprehensible discussion of these methods please refer to (Sebastiani, 2002) and the recent survey presented in (Aggarwal and Zhai, 2012). Below we introduce the data stream setting for extracting a model from data.

### 2.2. The Data Stream Setting

In this paper, we address text categorization under a *data stream setting*, i.e, we assume that documents arrive as a continuous and virtually infinite sequence of observations $d_1, d_2, \ldots, d_t, \ldots, , d_t \in \mathcal{D}$. In contrast to the traditional setting for learning a text classifier, the set of documents and the set of words (features) contained in those documents are both unknown beforehand, i.e., $D$ and $\mathcal{W}$ are being continuously updated. In agreement with recent papers on the topic (see e.g. Read et al., 2012; Mena Torres and Aguilar Ruiz, 2014) we recognize original requirements of systems devised to efficiently operate in this setting: (1) process an instance at a time and inspect it at most once; (2) be ready to predict at any point; (3) data may be evolving over time; and (4) expect an infinite stream, but process it under finite resources (time and memory).

An appropriate framework for learning from data under this setting is *online learning* (Pavlidis et al., 2011; Wang et al., 2012). Online learning takes place in a sequence of consecutive rounds $t = 1, 2, \ldots$. At each round $t$, the system is asked to predict the set of tags $\hat{T}_t \subset \mathcal{T}$ corresponding to a new observation $d_t \in \mathcal{D}$. To this end, the system stores a prediction mechanism $f_t : \mathcal{D} \to 2^{\mathcal{T}}$. Only after providing an output $\hat{T}_t = f_t(d_t)$, and if available, the system is feed with the set of correct labels $T_t$. Using this information, the learner can estimate its current performance and may activate a learning mechanism to get an improved decision mechanism $f_{t+1}$. In contrast to batch settings, in which the system knows all the correct labels in advance, an online system is being tested continuously as long as correct labels are provided. Each cycle corresponds to a testing and training step. Thus, online classification systems can be monitored by updating the so called *cumulative loss*, $L_T = \sum_{t=1}^{T} \ell(T_t, \hat{T}_t)$, where $\ell : \mathcal{T} \times \mathcal{T} \to \mathbb{R}$ is a loss function measuring the cost of a difference between $\hat{T}_t$ and $T_t$.

In section 5, we discuss some related work on the task of classifying data streams. Two important issues in adapting these techniques for document classification are the requirements of multi-label annotations and the large dimensionality arising from the traditional way to represent text.

*2.3. Text Representation*

Text representation is the task of transforming text documents into elements of a formally defined space $\mathcal{X}$ amenable for pattern analysis tasks. Because of its simplicity and effectiveness, the vector space model (VSM) is quite the most used text representation approach in document retrieval and text mining applications (Joachims, 2002; Zhang et al., 2011). In this model, text is represented as a vector $\mathbf{x} = (x_1, \ldots, x_d)$ whose components correspond to the weights of a set of linguistic units $\mathcal{V}$ used to capture the content of a document. Most frequently, $\mathcal{V}$ is constructed by extracting the words contained in the documents and the weights are determined by using count statistics ignoring both sequence and position in the text. For instance, given a document $d$ and word $w_i$ observed $F_i(d)$ times in the document, the TF representation of a text document can be obtained by setting $x_i = \text{TF}(w_i, d)$, where,

$$\text{TF}(w_i, d) = F_i(d)/F(d), \tag{1}$$

and $F(d)$ is a suitable normalization factor such as the document length in words. Usually, the word extraction process involves some pre-processing steps carried out to increase the quality of the representation. Typical pre-processing steps include (1) lower-casing, (2) removal of punctuation, (3) stop-word removal and (4) stemming (for details see e.g. Joachims, 2002). More sophisticated feature selection and dimensionality reduction methods can also be used to increase the quality of the vocabulary (Forman, 2003).

In addition to individual words, a number of authors have investigated the use of longer indexing units, linguistically or statistically meaningful for identification purposes, as they may have a smaller degree of ambiguity and become closer to expressing structured concepts (Zhang et al., 2011). Similarly, a number of more sophisticated weighting schemes have been investigated to enhance text classification (Lan et al., 2009). TF-IDF is probably the popular method (Joachims, 2002; Sebastiani, 2002). The basic idea is that a term which occurs in many documents is not a good discriminator and should be given less weight than one which occurs in few documents. This idea can be quantified using the so called inverse document frequency (IDF), which is commonly computed as

$$\text{IDF}(w_i) = \log(n/n_i), \tag{2}$$

where $n_i$ is the number of documents containing the term $w_i$. Combining IDF with the TF weight of Eqn.(1), leads to the TF-IDF weighting scheme

$$\text{TF-IDF}(w_i, d) = \text{TF}(w_i, d) \times \text{IDF}(w_i). \tag{3}$$

## 2.4. The Hashing Trick

The *hashing trick* is a method for speeding up the inner product computations required by some learning methods and improving their storage requirements. This was first presented by Shi et al. (2009b) and then extended in (Shi et al., 2009a) and (Attenberg et al., 2009). Here we briefly present the essential definitions and results that we require for the analysis of our technique.

Suppose data has been represented in a $d$-dimensional feature space $\mathcal{X}$ indexed by $[d] = \{1, 2, \ldots, d\}$ and let $h : [d] \to M := \{1, 2, \ldots, m\}$ with $M \ll d$ be a hash function drawn from a family of pairwise independent hash functions. Then, for a given vector $\mathbf{x} \in \mathcal{X}$, the hashing trick maps the original representation $\mathbf{x}$ to a new representation $\boldsymbol{\phi}(\mathbf{x})$ in the low dimensional space indexed by M. The coordinates of $\boldsymbol{\phi}(\mathbf{x})$ are defined as follows

$$\phi(\mathbf{x})_i := \sum_{j:h(j)=i} \xi(j) x_j, \qquad (4)$$

where $\xi : \mathbb{N} \to \{\pm 1\}$ is an auxiliary Rademacher hash function (Shi et al., 2009a). This "compressed" representation can then be used to approximate $\langle \mathbf{x}_1, \mathbf{x}_2 \rangle$ by computing

$$\langle \mathbf{x}_1, \mathbf{x}_2 \rangle_\phi := \langle \boldsymbol{\phi}(\mathbf{x}_1), \boldsymbol{\phi}(\mathbf{x}_2) \rangle. \qquad (5)$$

Approximating $\langle \mathbf{x}_1, \mathbf{x}_2 \rangle$ by (8) reduces the time required to compute a single inner product and the space required to store a single training instance from $\mathcal{O}(d)$ to $\mathcal{O}(m)$. As demonstrated by Shi et al. (2009a), $\langle \mathbf{x}_1, \mathbf{x}_2 \rangle_\phi$ is an unbiased estimator of $\langle \mathbf{x}_1, \mathbf{x}_2 \rangle$, i.e., $E_\phi(\langle \mathbf{x}_1, \mathbf{x}_2 \rangle_\phi) = \langle \mathbf{x}_1, \mathbf{x}_2 \rangle$. Furthermore, exponential tail bounds for the approximate preservation of norms and inner products were recently proved by Dasgupta et al. (2010). We state the main result below,

**Theorem 1.** *For any $\epsilon \in (0, 1)$, any $\delta < 0.1$ and any $\mathbf{x}$ satisfying the sparseness condition*

$$\eta^2 = \|\mathbf{x}\|_\infty^2 \leq \epsilon \log(1/\delta)^{-1} \log(m/\delta)^{-2}/16, \qquad (6)$$

*one has with probability at least $1 - 3\delta$ the following property*

$$(1 - \epsilon)\|\mathbf{x}\|_2^2 \leq \|\mathbf{x}\|_\phi^2 \leq (1 + \epsilon)\|\mathbf{x}\|_2^2, \qquad (7)$$

*provided $m \geq 12\epsilon^{-2} \log(1/\delta)$.*

Rephrasing, under mild conditions on the original representation space, the probability of a distortion in the inner-products $\langle \mathbf{x}_1, \mathbf{x}_2 \rangle_\phi$ with respect to the original inner product $\langle \mathbf{x}_1, \mathbf{x}_2 \rangle$ can be bounded as

$$\delta = \Pr\left(\frac{\langle \mathbf{x}_1, \mathbf{x}_2 \rangle_\gamma - \langle \mathbf{x}_1, \mathbf{x}_2 \rangle}{\langle \mathbf{x}_1, \mathbf{x}_2 \rangle} \geq \epsilon\right) \leq 6 \exp\left(-\frac{m\epsilon^2}{12}\right), \qquad (8)$$

that is, the probability of a large deviation of $\langle \mathbf{x}_1, \mathbf{x}_2 \rangle_\gamma$ from $\langle \mathbf{x}_1, \mathbf{x}_2 \rangle$ decays exponentially fast with the dimensionality $m$ of the embedding (4).

A similar result holds for the preservation of norms. A technique to obtain property (7) without explicitly requiring condition (6) was studied in (Weinberger et al., 2009). Essentially, this method pre-condition data by using a densification matrix $\mathbf{P}$. This increases the computational cost of computing the embedding by a factor of $c = \mathcal{O}(1/\epsilon)\mathcal{O}(\log(1/\delta))$, but leads to a result valid for any type of data.

*2.5. The Count-Min (CM) Sketch*

The hashing trick presented above has its roots in the so-called *count-min sketch* proposed by Cormode and Muthukrishnan (2005); a probabilistic data structure designed to efficiently track the observed frequency distribution $f(\cdot)$ of a stream of elements $D$ from a domain $N$. We will use this technique to obtain an online approximation of the IDF weighting scheme.

Let $f(j)$ denote the number of elements in $D$ having a value $j \in N$. A count-min sketch to approximate $f(j)$ within a precision $\epsilon$ and a confidence $\delta$, is based a two-dimensional array $C$ of $L \times m$ counters. For each row $\ell$, a hash function $h_\ell(\cdot)$ maps the input domain $N$ uniformly into the range $M = \{1, 2, \ldots, m\}$. If, at a given time, element $j$ is observed $m_j$ times, each row of the sketch is updated as follows

$$C_{\ell,h_\ell(j)} = f(h_\ell(j)) + m_j, \qquad (9)$$

that is, the counts for the element $j$ are stored in the position $h_\ell(j)$ of each row $\ell$. Of course, position $h_\ell(j)$ may also store the counts for other elements $k$ such that $h_\ell(k) = h_\ell(j)$. If, at a given time, an estimation of $f(j)$ is required, the sketch outputs

$$\hat{f}(j) = \min_\ell(C_{\ell,h_\ell(j)}). \qquad (10)$$

Cormode and Muthukrishnan (2005) proved the following theorem about the the estimate $\hat{f}(j)$.

**Theorem 2.** *Given parameters $\epsilon > 0, \delta \in (0, 1)$, set $m = \lceil \exp(1)/\epsilon \rceil$ and $L = \lceil \ln(1/\delta) \rceil$. Thus, the estimate $\hat{f}(j)$ in Eqn.(10) satisfies: $f(j) \leq \hat{f}(j)$, and with probability at least $1 - \delta$,*

$$\hat{f}(j) \leq f(j) + \epsilon \|\mathbf{f}\|_1, \qquad (11)$$

where $\|\mathbf{f}\|_1 = \sum_j f(j)$.

## 3. THE CLASHING SYSTEM

A clashing system is composed of the following main components: (1) a document wise *hashing function* $\phi : \mathcal{D} \to \mathbb{R}^m$, used to get a low-dimensional vector representation of a document from the set of terms (words) it contains; (2) a *partition* $\mathcal{R} = \{\mathcal{R}_1, \ldots, \mathcal{R}_{n_p}\}$ of the low-dimensional feature space $\phi(\mathcal{D})$; (3) a *map function*, used to automatically accommodate a document in a region of the partition; (4) a *prediction rule* used to output a set of labels according to the region; (5) a *learning rule*, used to update the prediction rule from labelled instances. Algorithm 1 depicts the general operation of the system.

In a nutshell, the principle underlying clashing is that similar documents will collide in the same region with high probability and thus classification can be performed by storing a set of simple statistics about the label distribution. In the next paragraphs we present each component in detail.

**Algorithm 1:** Overview of the Clashing System.

**Input**: A stream of documents $d_1, d_2, \ldots$ with tag sets $T_1, T_2, \ldots$.
1. Initialize each component of the system.
2. **for** $d_1, d_2, \ldots$ **do**
3.     Hash the document: $\boldsymbol{\phi}(d_t) \leftarrow \texttt{doc-hash}(d_t)$.
4.     Map to a region: $j_* \leftarrow \texttt{map}\,(\boldsymbol{\phi}(d_t))$.
5.     Predict labels: $T_* \leftarrow \texttt{predict}\,(j_*)$.
6.     Update the prediction mechanism: If $T_i \neq \emptyset$, $\texttt{learn}\,(d_t, T_*, T_i)$.

### 3.1. Document Hashing

We have shown in the previous section that any vector in a metric space $\mathcal{X}$ can be quickly embedded in a low-dimensional space $\boldsymbol{\phi}(\mathcal{D})$ such that the norms and inner products are approximately preserved. Thus, if $\mathbf{x}(d)$ represents the TF representation of $d$, the hashed representation $\phi(\mathbf{x})$ in Eqn.(4) approximately preserves the geometry of the word count space. A direct application of this "hashing trick" in a batch setting would thus proceed as follows:

1. Create a vocabulary $\mathcal{V} = \{w_1, w_2, \ldots\}$, containing the words in the corpus, and assign an arbitrary index $j \in \mathbb{N}$ to each of them.
2. Build the TF representation $\mathbf{x}(d)$ for each document $d$ using $\mathcal{V}$ and Eqn. (1).
3. Use the hashing map of Eqn.(4) with $d = |\mathcal{V}|$ to reduce dimensionality.

However, in data stream scenarios the vocabulary $\mathcal{V}$ is unknown in advance and is indeed continuously growing. An important advantage of the hashing method is that $\phi(\mathbf{x})$ can be computed by simply scanning the words in $d$ one by one, without explicitly building the vocabulary. This procedure is presented as Algorithm 2.

**Algorithm 2:** $\texttt{doc-Hash}(d)$ for preserving the TF geometry.

**Input**: A document $d$ with words $w_1, w_2, \ldots$
**Output**: $\phi(\mathbf{x})$ where $\mathbf{x}$ is the TF representation of $d$.
**Initialization**: No required.
1. $\boldsymbol{\phi}(d) \leftarrow \mathbf{0}$
2. **for** *Each word $w_i$ in the document* **do**
3.     $k \leftarrow h(w_i)$.
4.     $\boldsymbol{\phi}(d)_k \leftarrow \boldsymbol{\phi}(d)_k + \xi(w_i)$.
5. Output $\boldsymbol{\phi}(d)$.

Note that here, the hashing functions $h : \mathcal{W} \to M := \{1, 2, \ldots, m\}$ and $\xi : \mathcal{W} \to \{\pm 1\}$ operate directly on strings. It is easy to see that after the loop in Algorithm 2, to process all the words $w_1, w_2, \ldots$ in a document $d$, the hashed representation $\boldsymbol{\phi}(d)$ has components $\boldsymbol{\phi}(d)_i = \sum_{w:h(w)=i} \xi(w)\mathbf{x}(w)$, where $\mathbf{x}(w)$ is the number of occurrences of the word $w$ in the document, i.e, the $j$-th coordinate in the TF representation of $d$, assuming that $j$ is the index assigned to the word $w$ in the vocabulary. Since the indexing in $\mathcal{V}$ is arbitrary, applying Algorithm 2 to each document in the corpus is equivalent to performing the batch steps.

As we have previously discussed, a term weighting scheme $\boldsymbol{\omega} : \mathcal{V} \to \mathbb{R}_0^+$ is usually performed on the TF representation in order to enhance classification performance. In particular, IDF weights have demonstrated to be highly effective in practice (Sebastiani, 2002; Joachims, 2002). Here we provide an efficient online method to preserve the geometry of the word count space corresponding to the TF-IDF representation. In batch scenarios we may explicitly scale the TF representation of documents, using an IDF dictionary computed in an offline fashion from the corpus, and then applying the hashed feature map of Eqn.(4) to the obtained feature vector in order to reduce dimensionality. However, in online scenarios, this approach forces a periodic recomputation of the IDF weights for a probably large vocabulary $\mathcal{V}$. Algorithm (3) provides a method to perform this computation online.

---

**Algorithm 3:** `doc-Hash`$(d)$ for preserving the TF-IDF geometry.

**Input**: A document $d$.
**Output**: A low dimensional approximation to the TF-IDF representation of $d$ (counters $C \in \mathbb{R}^m, n \in \mathbb{R}$ are updated and stored for the next round).
**Initialization**: $n \leftarrow 0$, $C \leftarrow \mathbf{0}$

1   $\hat{\boldsymbol{\phi}}(d) \leftarrow \mathbf{0}$, $\bar{C} \leftarrow \mathbf{0}$.
2   **for** *Each word $w_i$ in the document* **do**
3      $k \leftarrow h(w_i)$.
4      $\hat{\boldsymbol{\phi}}(d)_k \leftarrow \hat{\boldsymbol{\phi}}(d)_k + \xi(w_i)$.
5      **If** $\bar{C}_k = 0$, **then** $\bar{C}_k \leftarrow 1$.
6   $C \leftarrow C + \bar{C}$.
7   $n \leftarrow n + 1$.
8   **for** *Each component $k$ in $\hat{\boldsymbol{\phi}}(d)_k$* **do**
9      $\overline{\mathrm{IDF}}(k) \leftarrow \log(n/C_k)$
10     $\hat{\boldsymbol{\phi}}(d)_k \leftarrow \overline{\mathrm{IDF}}(k)\hat{\boldsymbol{\phi}}(d)_k$.
11 Output $\boldsymbol{\phi}(d) := \hat{\boldsymbol{\phi}}(d)$.

---

Essentially, Algorithm 3 differs from Algorithm 2 in the scaling performed in step 10. This scaling is aimed to approximate the IDF weighting scheme and it is performed by storing a hashed representation of the vector $\mathbf{f}$ containing the true document frequencies of words among the $n$ documents observed up to a given round. This approximation is represented by the array $C$. Indeed, is not hard to see that after observing $n$ documents, $C$ has components $C_k = \sum_{j:h(i)=k} f(i)$, where $f(i)$ is the document frequency of the word $w_i$ among the $n$ documents, that is, the number of documents containing at least once the term. Thus, array $C$ in Alg.3 is a direct implementation, with $L = 1$, of the method presented by Cormode and Muthukrishnan (2005) to approximate $f(i)$ and thus enjoys the properties given in Theorem 2. For a given $w_i$, the exact IDF computed on set of documents observed by the system would proceed by computing $\mathrm{IDF}(w_i) \leftarrow \log(n/f(i))$. Our method, instead, computes $\overline{\mathrm{IDF}}(k) \leftarrow \log(n/C_k)$ for any word colliding in the $k$-th coordinate of $C$. In the next section, we use the bound of Eqn.(11) to bound the probability that $\log(n/C_k)$ be different than the true IDF weight $\log(n/f(i))$ of any word $w_i$ such that $h((w_i) = k$.

---
**Algorithm 4:** (The Mapper) `map(d)`.
---
**Input**: A document $d$ represented as $\phi(d)$.
**Output**: The index $j_*$ of the region for clashing $d$.
**Initialization**: No required.
1 Find the Nearest Prototype: $j_* = \arg\max_i s(\phi(d), \mathbf{p}_i)$.
2 Output $j_*$.
---

*3.2. The Mapper*

The classification model underlying the clashing system can be represented by a partition $\mathcal{R} = \{\mathcal{R}_1, \ldots, \mathcal{R}_{n_p}\}$ of the low-dimensional space $\phi(\mathcal{D})$ in which documents are embed. A function called Mapper is aimed to actively make similar documents to collide in the same regions. A simple way to achieve this goal is by using a set of prototypes $\mathbf{p}_1, \mathbf{p}_2, \ldots, \mathbf{p}_{n_p} \in \phi(\mathcal{D})$ and a Voronoi rule of the form

$$\mathcal{R}_i := \{\phi(d) : i = \arg\min_j \|\phi(d) - \mathbf{p}_j\|^2\}. \tag{12}$$

By Eqn.(12), the documents assigned to the same regions have the same nearest prototype. If denote by $2\eta$ the maximum distance from a point in $\phi(\mathcal{D})$ to its corresponding prototype (the extreme points of $\mathcal{R}_i$), we have that two documents $d_1, d_2$ assigned to the same region are similar by at least $s(\phi(d_1), \phi(d_2)) \geq \eta$. Since $\phi(d)$ preserves similarity with a high probability, $d_1, d_2$ are likely to be close in the TF or TF-IDF space. The Mapper is then be implemented as in Alg. 4.

*3.3. The Prediction Rule*

Each region $\mathcal{R}_i$ is associated with a probability distribution on $\mathcal{T}$, denoted $F_i = \{F_{i,1}, \ldots, F_{i,n_t}\}$, which models the probability of observing a given tag $\tau_j$ in a document assigned to that region, that is, $F_{i,j} = \Pr(\tau_j | \mathcal{R}_i)$ or better, $F_{i,j} = \Pr(\tau_j \in T(d) | \phi(d) \in \mathcal{R}_i)$, where $T(d)$ denotes the set of tags for a document $d$. Formally, the classification hypothesis $f : \mathcal{D} \to 2^{\mathcal{T}}$ is implemented as

$$f(d) = \{\tau_i \in \mathcal{T} : \hat{F}_{i,j_*} > \theta\}, \tag{13}$$

where $\hat{F}_{i,j}$ is an estimator of $F_{i,j}$ and $\theta$ is a parameter that may be used to control the precision/recall tradeoff of the system. That is, a label is predicted by the system if and only if a fraction at least $\theta$ of the documents colliding in the region $j_*$ contains that tag. We adopt $\theta = 0.5$ in this paper.

*3.4. The Learning Engine*

The learning engine is the component of the system determining the way to construct a suitable partition and estimate the label distributions from data. Considering the decisions made before, the first question translates into how to construct the prototypes from a set of examples $D = \{(d_1, T_1), \ldots (d_n, T_n)\}$. As to keep the model simple and efficient, we consider linear transformations of subsets of training data $\phi(D) = \{\phi(d_1), \ldots, \phi(d_n)\}$, i.e., if $\phi\mathbf{D}$ denotes the matrix with the examples arranged in the columns, $\mathbf{p}_i := \phi\mathbf{D}\boldsymbol{\alpha}$, where $\boldsymbol{\alpha}$ is a set of parameters. In addition, we restrict our analysis to recursive update rules of the form

$$\mathbf{p}_i^{t+1} = (1 - \lambda_i^t)\mathbf{p}_i^t + \lambda_i^t \phi(d_t), \tag{14}$$

where $d_t$ is the $t$-th document observed by the system. Parameter $\lambda_i^t \in [0,1]$ allows us to track the set of examples colliding together in region $\mathcal{R}_i$. Different schemes to automatically set $\lambda_i^t \in [0,1]$ from empirical data can be devised. In this paper, we restrict the analysis to the following simple approaches:

- **Mode 1**. We associate one prototype $\mathbf{p}_i$ to each class label $\tau_i$ and compute $\mathbf{p}_i$ as the centroid of all the documents containing this tag. That is, at a given round $t$, $\lambda_i^t = 0$ if and only if $T_t$ contains $\tau_i$. Otherwise $\lambda_i^t = 1/(n_i^t + 1)$, where $n_i$ is the number of documents containing $\tau_i$ up to round $t$. Algorithm 5 summarizes this procedure.

- **Mode 2**. We associate one prototype $\mathbf{p}_i$ to each class label $\tau_i$ but compute $\mathbf{p}_i$ only from the examples leading to a classification mistake. This is essentially the principle behind the perceptron rule. Here we devise a multi-label version. Given a training document $d_t$ with labels $T_t$, the system is trained as follows. If a label $\tau_i \in T_t$ is not predicted by the system (false negative), we feed the example to $\mathbf{p}_i$. In this case, $\lambda_i^t = 1/(n_i^t + 1)$. This searches to increase the recall of the system. If one label $\tau_i \in T_t$ is incorrectly assigned to $d_t$, we apply the rule explained in the item 1. Algorithm 6 summarizes this learning rule.

To estimate the probabilities $F_i$, we adopt a frequentist approach, computing

$$\hat{F}_{i,j} := \frac{|\phi(D)_i \cap \phi(D)_j|}{|\phi(D)_i|} = \frac{n_{i,j}}{n_i}, \tag{15}$$

where $n_{i,j}$ is the number of training documents containing tags $\tau_i$ and $\tau_j$. This criterion can be implemented in online learning using the following simple recursion

$$\hat{F}_{i,j}^{t+1} = \begin{cases} (1-\lambda_i^t)\hat{F}_{i,j}^t + \lambda_i^t & \text{if } \tau_i \in T_t \text{ and } \tau_j \in T_t \\ (1-\lambda_i^t)\hat{F}_{i,j}^t & \text{otherwise} \end{cases} \tag{16}$$

Note that setting $\lambda_i^t = 1/n_i^t$ easily leads to Eqn.(15). Definition of Eqn. (16) may be used to set more general schemes to adapt $\hat{F}_{i,j}$ from labelled examples. Algorithms 5 and 6 incorporate the estimation of the local statistics.

---

**Algorithm 5:** The Learning Engine - Mode 1.

**Input**: $\phi(d_t)$, predicted tags $T_*$, and correct tags $T_t$.
**Output**: Updated partition $\mathcal{R}$ and local statistics $F_i$.
**Initialization**: Create one prototype for each class, setting $\mathbf{p}_i \leftarrow \mathbf{0}, \forall i$, $\hat{F}_{i,j} \leftarrow 0, \forall i,j$

1 **for** *Each label $\tau_i$ contained in $T_t$* **do**
2      Set $n_i \leftarrow n_i + 1$ and $\lambda_i \leftarrow 1/n_i$
3      Update $\mathbf{p}_i \leftarrow (1-\lambda_i)\mathbf{p}_i + \lambda_i \phi(d_t)$
4      **for** *Each $\hat{F}_{i,j}$ in $\hat{F}_i$* **do**
5          If $\tau_j \in T_t$, $\hat{F}_{i,j} \leftarrow (1-\lambda_i)\hat{F}_{i,j} + \lambda_i$ // *Local Statistics*
6          Else $\hat{F}_{i,j} \leftarrow (1-\lambda_i)\hat{F}_{i,j}$

**Algorithm 6:** The Learning Engine - Mode 2.

**Input**: $\phi(d_t)$, predicted tags $T_*$, and correct tags $T_t$.
**Output**: Updated partition $\mathcal{R}$ and local statistics $F_i$.
**Initialization**: Create one prototype for each class, setting $\mathbf{p}_i \leftarrow \mathbf{0}, \forall i, \hat{F}_{i,j} \leftarrow 0, \forall i,j$

1. $F_n = T_t - T_*$ // *False Negatives Labels*
2. $F_p = T_* - T_t$ // *False Positive Labels*
3. **for** *Each label $\tau_i$ contained in $F_n$* **do**
4.     Set $n_i \leftarrow n_i + 1$ and $\lambda_i \leftarrow 1/n_i$
5.     Update $\mathbf{p}_i \leftarrow (1 - \lambda_i^t)\mathbf{p}_i + \lambda_i^t \phi(d_t)$
6.     **for** *Each $\hat{F}_{i,j}$ in $\hat{F}_i$* **do**
7.         If $\tau_j \in T_t$, $\hat{F}_{i,j} \leftarrow (1 - \lambda_i)\hat{F}_{i,j} + \lambda_i$ // *Local Statistics*
8.         Else $\hat{F}_{i,j} \leftarrow (1 - \lambda_i)\hat{F}_{i,j}$

9. If $F_p \neq \emptyset$, call Learning Engine - Mode 1.

## 3.5. Complexities

The system needs to store a fixed set of $n_p$ prototypes. In this paper, $n_p = n_t$, that is, we have one prototype for each possible label. Classification and training require a search among the prototypes, using a similarity function linear in the compressed space dimensionality. Thanks to the hashing based embedding, the space and running time complexities are independent of the potential number of attributes (words in text). Space complexity amounts to $\mathcal{O}(n_p m) + \mathcal{O}(n_p n_t)$ because we store $n_p$ $m$-dimensional prototypes and $n_t$ counters for each region. Test complexity amounts to $\mathcal{O}(n_p m)$ due to the nearest prototype search. Update time is $\mathcal{O}(|T_t|(n_p m + n_t))$ where $|T_t|$ is the number of labels contained in the document $d_t$. Note that both storage and running time are independent of the number of observed documents. Parallelization of prototypes storage and search (in applications with a huge number of labels) is straightforward.

## 3.6. Beyond TF-IDF

Considering the increasing interest on weighting methods capable to improve the accuracy of text classifiers in the last years (Lertnattee and Theeramunkong, 2004; Lan et al., 2009; Guan et al., 2009; Altincay and Erenel, 2010; Luo et al., 2011; Erenel and Altincay, 2012; Ren and Sohrab, 2013; Wang and Zhang, 2013), it may be relevant to ask if the online method provided here to approximate TF-IDF can be extended to approximate other weighting schemas. For instance, several papers have recently investigated methods capable to exploit information about the different distribution of terms among documents of each class. Other works have studied variants of TF-IDF based on feature selection scores (filter methods) originally devised to reduce dimensionality (Azam and Yao, 2012; Sebastiani, 2002; Lan et al., 2009; Altincay and Erenel, 2010).

Our online embedding procedure is based on two probabilistic methods: one for approximating TF and one for approximating IDF. These two components interact as depicted in Algorithm 3. The guarantees for the probably approximately correct approximation of IDF are based on the properties of the count-min sketch presented in section 2.5. Using this data structure, we can approximate the document frequency $\sum_t I(t_i \in d_t)$ (DF) of a term $t_i$ in the stream with high accuracy. However, the transformation applied on a DF estimate to obtain an IDF estimate has an impact on the final accuracy of the approximation. Similarly, our guarantees for the approximately correct computation of TF are based on the properties of the hashing map studied in section 2.4. Functions of TF computed on our estimator can be more or less accurate depending on the specific transformation involved. If the weighting method is based on a "simple" transformation of DF, IDF and/or TF, as defined in our manuscript, an extension of our approach is likely to be painless. However, if the transformation applied by the term weighting method is more "complex", for instance it is not continuous, it is highly non linear and/or dependent on several other counts that we are not currently computing/estimating, a more acute analysis is required. Specifically we would need to check if Lemma 1 and Proposition 1 can be adapted for the specific weighting scheme at hand.

In order to provide a more specific answer to the question, we can focus on some specific approaches. Consider for instance the class-based weighting method proposed by Ren and Sohrab (2013), which consistently outperformed other term weighting approaches using SVM and centroid-based classifiers. This method works by incorporating in TF-IDF a novel component which depends on the class conditional document frequency $\sum_t I(t_i \in d_t \wedge d_t \in C_k)$ (let us say CDF) of a term $t_i$ in the corpus. The sketch we use to approximate DF can be easily adapted to approximate these class conditional document frequencies from a data stream. It would be enough to store $K$ sketches $C^1, C^2, \ldots, C^K$ as those used by Algorithm 3, one for each category in the problem, and update $C^j$ from the flow of terms coming from documents containing label $j$. Therefore, if we can accept an slight additional cost in terms of storage and computation, our method may be adapted to approximate CDF with about the same accuracy as we currently approximate DF. The precision obtained for the final weighting scheme depends on the function mapping CDF to weights.

The first scheme proposed in (Ren and Sohrab, 2013) weights the standard TF-IDF using a component defined in terms of the average of the class-conditional document frequencies,

$$w(t_i) = 1 + \log\left(\frac{C}{CS_\delta(t_i)}\right) , \quad CS_\delta(t_i) = \sum_{c_k} \frac{n_{ck}(t_i)}{N_{c_k}} ,$$

where $n_{ck}(t_i)$ denotes the number of documents that include the term $t_i$ and are a member of the category $c_k$, and $N_{c_k}$ denotes the total number of documents in a certain category $c_k$. Since $c_k$ coincides with our definition of CDF, the sum in $CS_\delta(t_i)$ is a convex function, and the transformation $\omega(\cdot) = 1 + \log(C/\cdot)$ employed in this scheme is essentially the same that the transformation applied on DF to compute IDF, it is easy to adapt Lemma 1 and Proposition 1 in order to provide guarantees for the new weighting schema.

The second approach proposed in Ren and Sohrab (2013) is based on Inverse Category Frequency, an idea explored also in (Wang and Zhang, 2013; Lertnattee and Theeramunkong, 2006) and (Lertnattee and Theeramunkong, 2004). It scales TF-IDF by the additional factor

$$\omega(t_i) = 1 + \log\left(\frac{C}{c(t_i)}\right) ,$$

where $C$ is the number of classes and $c(t_i)$ is the number of categories in which the term $t_i$ occurs at least once. It is clear that $C$ can be computed exactly. In order to approximate $c(t_i)$ for each possible $t_i$, a straightforward method is to rely on the $K$ Count-Min Sketches $C^1, C^2, \ldots, C^K$ we have described above. Given the hashing function $h(\cdot)$ and a term $t_i$ such that $h(t_i) = j$, $c(t_i)$ may be approximated as $\bar{c}(t_i) = \sum_j I(C_j^k)$, where $I(x)$ is 1 if $x > 0$ and 0 otherwise. Since $C_j^k$ approximates $n_{ck}(t_i)$, $\bar{c}(t_i)$ should approximate $c(t_i)$ reasonably well. However, in order to preserve the guarantees provided by Lemma 1 and Proposition 1, we would need to examine the effect of the discontinuity in the function $I(x)$. Clearly, we should also focus on designing a more efficient sketch to directly approximate $c(t_i)$ without storing $K$ different data structures $C^1, C^2, \ldots, C^K$.

Another approach for more accurate term weighting in text categorization is based on feature selection metrics such as chi-squared, Gini, information gain, odds ratio, etc. (Azam and Yao, 2012; Lan et al., 2009; Altincay and Erenel, 2010). Given a term $t_i$ these metrics can be defined using the following counts

1. $A_k[t_i]$: number of documents that contain $t_i$ and belong to class $k$.
2. $B_k[t_i]$: number of documents that do not contain $t_i$ and belong to class $k$.
3. $C_k[t_i]$: number of documents that contain $t_i$ and do not belong to class $k$.
4. $D_k[t_i]$: number of documents that do not contain $t_i$ and do not belong to class $k$.

Clearly, $A_k[t_i]$ coincides with our definition of CDF for class $j$ and term $t_i$. In addition, $B_j[t_i] = N_{c_k} - A_k[t_i]$. Therefore, it is not hard to see from our previous discussion that we can adapt our method to approximate $A_k[t_i]$ and $B_k[t_i]$ with high accuracy at a slight increase in computational cost. On the other hand, $C_k[t_i] = \sum_t I(t_i \in d_t \wedge d_t \in C_k) - A_k[t_i]$. Both terms in the subtraction can be accurately approximated. Finally, $D_k[t_i] = N - A_k[t_i] - B_k[t_i] - C_k[t_i]$. Thus, we can approximate all these counters using the $K$ class-based sketches described above. This is a good start for an extension of our method to weighting schemes using feature selection metrics. However, metrics can depend on $A_k[t_i], B_k[t_i], C_k[t_i], D_k[t_i]$ in very complex way (see e.g. Forman, 2003; Lan et al., 2009). Therefore, we would need to proceed case by case, considering the relevant metrics and studying a way to produce accurate final estimations.

## 4. THEORETICAL ANALYSIS

We present an analysis of the online embedding procedure used by the clasher aimed to theoretically characterize the effect of the dimensionality reduction on the classification performance. We compare the predictive performance of the model in the hash space $\phi(\mathcal{D})$ to the performance of the equivalent model in the word count space $\mathcal{X}$. If $\mathcal{X}$ denotes the TF representation of the document space, we know from the discussion in the previous section that $\phi(\mathcal{X})$ approximately preserves the geometry (norms and inner products) of the space $\mathcal{X}$. This is a direct consequence of Theorem 1. However if $\mathcal{X}$ denotes the TF-IDF representation of the document space, Theorem 1 does not suffice because we need to account for the effect of the approximate IDF weighting used in Alg. 3.

### 4.1. Preservation of the TF-IDF Geometry

Our first result is that approximate IDF weights computed in Alg. 3 enjoy essentially the same guarantees that Theorem 2 provides for the sketch $C$. i.e., the operation $\log(n/\cdot)$ does not essentially change the theoretical bounds on the difference between the true and the estimated quantities. The proof of this lemma is provided in the appendix.

**Lemma 1.** *At a given round $t$, let $IDF(w_i)$ be the exact IDF for a word $w_i$ computed from the documents observed so far, and $h(w_i) = k$. Thus, with constant probability, the IDF estimate $\overline{IDF}(w_i) = \log(n/C_k)$ computed by Alg. 3 satisfies $\overline{IDF}(w_i) \leq IDF(w_i)$ and $\overline{IDF}(w_i) \geq IDF(w_i) - \epsilon \|\mathbf{f}\|$, for any $\epsilon > 0$, provided $m \geq \lceil \exp(1)/\epsilon \rceil$. Here, $\|\mathbf{f}\|$ is defined as in Theorem 2, i.e., it corresponds to the sum of all the document frequencies.*

The next proposition puts together the guarantees about the approximate IDF weights and the guarantees for the hashed TF representation, in order to show that Alg. 3 computes a mapping that approximately preserves the exact TF-IDF geometry for any desired precision $\bar{\epsilon}$ provided the dimensionality of the hash function is large enough. The proof of this proposition is provided in the appendix.

**Proposition 1.** *Under the conditions of Theorem 1, the representation $\phi(d_t)$ computed by Alg. 3 at round $t$ approximates the exact TF-IDF representation $\mathbf{x}_t$ of $d_t$, i.e., we have with probability at least $(1 - e^{-1})(1 - 3\delta)$ that the following property holds*

$$(1 - \bar{\epsilon})\|\mathbf{x}_t\|_2^2 \leq \|\phi(\mathbf{x})\|^2 \leq (1 + \bar{\epsilon}^2)\|\mathbf{x}_t\|_2^2, \tag{17}$$

*provided $m \geq \max\left(588\|\mathbf{f}\|^4 \bar{\epsilon}^{-2} \log(1/\delta), \lceil 7\|\mathbf{f}\|^2 \exp(1)/\bar{\epsilon} \rceil\right)$, for any $\bar{\epsilon} \in (0, 1)$ and any $\delta < 0.1$.*

### 4.2. Effect on the Classification Performance

We analyze the cumulated performance of the system at a given instant $t$ compared to the performance of an equivalent system operating directly in the word-count space. Thanks to the Theorem 1, the word count space can correspond to either the TF representation or the TF-IDF representation, with slightly different conditions on the dimensionality required to obtain the guarantees. For simplicity we analyze the case the TF representation using Theorem 1. For the TF-IDF the procedure is analogous using Theorem 1. For brevity we also assume that the clashing system is operating with the learning rule of Alg. 5.

The following result shows that the chance of achieving the same performance depends on the following notion of *margin*

$$\eta(\mathbf{x}) = \min_{j \neq i*} \|\bar{\mathbf{p}}_j - \mathbf{x}\|^2 - \|\bar{\mathbf{p}}_{i*} - \mathbf{x}\|^2 \tag{18}$$

**Lemma 2.** *Suppose that $\mathbf{x}$ is classified with margin at least $\eta$ in the original data space. For any $\bar{\delta} < 0.1$, set $\bar{\eta} \leq \eta/\max_j \|\bar{\mathbf{p}}_j - \mathbf{x}\|^2$, $m \geq 48\bar{\eta}^{-2} \log(3n_t/\bar{\delta})$ and $c = 32\bar{\eta}^{-1} \log(3n_t/\bar{\delta}) \log^2(3mn_t/\bar{\delta})$. Thus, if $\mathbf{x}$ satisfies $\|\mathbf{x}\|_\infty \leq 1/\sqrt{c}$, the probability that the hash embedding preserves the original decision is at least $1 - \bar{\delta}$.*

The proof of the latter lemma and the next proposition are provided in the appendix.

**Proposition 2.** *Let $D^t = \{d_1, d_2, \ldots d_{n_t}\}$ be a test set such that $\mathbf{x}(D^t) = \{\mathbf{x}_1^t, \ldots, \mathbf{x}_{n_t}^t\}$ is contained in a ball of diameter $D \in \mathbb{R}$. Let $Pr(Err)$ be the test error incurred by the clasher in the hash space and $Pr(\overline{Err})$ the test error in the original data space. For any $\eta* > 0$ and $\bar{\delta} < 0.1$, set $\bar{\eta} = \eta*/D^2$ and keep the parameters as in Lemma 2. Therefore, if $\forall k \in [n_t], \|\mathbf{x}_k^t\|_\infty \leq 1/\sqrt{c}$, we have*

$$Pr(Err) \leq Pr(\overline{Err}) + (1 - \delta_{\eta*})\bar{\delta} + \delta_{\eta*}, \tag{19}$$

where $\delta_{\eta*}$ is the fraction of test instances classified with margin $\eta(\mathbf{x}) < \eta*$. Indeed, if all the data $\phi(D) = \{\mathbf{x}_1, \ldots, \mathbf{x}_{n_t}\}$ is correctly classified with margin $\eta* > 0$ in the original space,

$$Pr(Err) \leq \bar{\delta} \leq 3n_t \exp\left(-\frac{m\eta*^2}{48D^4}\right). \qquad (20)$$

## 5. ADDITIONAL RELATED WORK

In this paper, we have been interested in constructing a very efficient system to classify documents coming from a large and high-speed data stream using bounded resources for every training and testing round. Therefore, this work intersects several issues in the field of data mining: data streams, text representation and multi-label classification. As we have previously discussed in section 1, few works approach text representation and multi-label classification in data stream settings. On the other hand, our solution to the problem relates to recent advances in data stream sketching, hashing and centroid based text classification. Here, we combine these ideas in a novel and original way to solve the problem at hand. In this section, we comment some additional related work to the proposal.

*Data Stream Methods.* Efficient data stream classification is currently approached in one of the following two ways. One is enabling a traditional machine learning technique with an online or incremental learning rule which allows the method to continuously learn from new observations. The VFDT (very fast decision tree) proposed by Domingos and Hulten (2000) is one of the most popular algorithms in this category. It extracts and keeps an anytime decision tree classifier using bounded computational resources with a performance similar to that of a batch implementation. Several extensions to deal with continuous attributes has been proposed (see e.g. Gama et al., 2003). Another example of this approach is the family of methods based on the perceptron algorithm presented by Crammer et al. (2006) to approximate SVMs in online learning settings. Law and Zaniolo (2005) presented an adaptation of the nearest neighbor algorithm to adaptively determine a suitable neighborhood in single-label data stream scenarios. The other mainstream approach for data stream classification consists in decomposing the stream into batches, training a different classifier on each batch and using a voting scheme to implement an ensemble decision function. The method developed by Oza (2005) is a baseline in this category. It extends bagging for online data streams and accepts any type of online classifier as base learner. An online version of boosting is also proposed but it is significantly more expensive computationally. Recently Bifet et al. (2009) has extended the work of Oza (2005) to deal with concept drift by designing a change detector to decide when to discard underperforming learners. This provides a general way to extend models supporting online learning (such as those studied in this paper) to drifting environments.

*Multi-label Classifiers.* Methods allowing multi-label text annotations have largely relied on the Binary Relevance (BR) decomposition method and its variants. Multi-label classification is achieved by training several binary classifiers on atomic tags, using the documents containing them as positive examples and the rest a negative examples. An advantage of this approach is that it makes possible to employ accurate techniques, such as support vector machines (SVMs), to solve the obtained sub-problems. Applying BR to data streams is straighforward by adopting online or incremental binary base models (see for instance Chai et al., 2002). The BR approach however has been largely criticized in the literature because it fails to take into account possible correlations among labels during the classification process. On the other hand, the obtained sub problems tend to be highly assymetric and it has been observed as a consequence that the approach usually overwhelm the class with more samples. Following Tsoumakas et al. (2010), BR can be classified as a problem transformation method where the representation of the dataset is changed in such a way that single-label classification methods can be directly applied. Other methods in this category are those using binary classifiers to predict label sets instead of atomic labels Read et al. (2011), or the pairwise classification method, where binary models are used for every possible pair of labels. Several adaptation methods where multi-label data is modeled at once has also been investigated, including extensions of boosting Schapire and Singer (2000) and neural networks Zhang and Zhou (2006). Most of this work focus on classic test/train settings and thus it is not well known how these approaches can work in data stream environments. A important contribution to provide baselines in this context is the work of Read et al. (2012) where several multi label methods are studied under a data stream settings. Here the authors propose an extension of the VFDT method Domingos and Hulten (2000) to handle multi-label predictions by incorporating multi-label classifiers at the leaves. This method was also combined with the ensemble methods of Bifet et al. (2009) and Oza (2005) in order to handle concept drift.

*Low Dimensional Text Representation.* The online text representation method we investigate in this paper is based on the technique presented by Shi et al. (2009b) to deal with high-dimensional data. Theorem (1) essentially shows that the hashing based feature map of Eqn. (4) can be used to implement the Johnson-Lindenstrauss (JL) lemma (Matoušek, 2008). In the last years, random projections (RP) (Achlioptas, 2003) have emerged as a popular way to achieve the same goal, leading to an efficient dimensionality reduction approach that has been used to attack several data mining problems. A number of methods to improve the computational cost of performing RP have been recently investigated, leading to the so called *fast Johnson-Lindenstrauss transforms* (Ailon and Chazelle, 2009). We can identify two key advantages in using a hashing based representation instead of RP for data stream applications. The first is that RP needs to know in advance the dimensionality of the original space. In some applications, such as those involving text, constraining in advance the set of potential attributes (e.g. words in text) prevent the system to fully exploit new data. The second is that RP needs to explicitly store the projection matrices that lead, via inner product computations, to the compressed representation. This leads to an additional storage cost and prevents to enjoy truly constant memory requirements in applications where the number of features is allowed to grow. In (Shi et al., 2009a), the idea of using a low dimensional sketch (Cormode and Muthukrishnan, 2005) to approximate the TF-IDF representation was applied to large-scale corpora but it was not explored in data stream settings. Recently, Baena-Garcia et al. (2011) extended this method to allow efficient representation of massive streams of documents but the effects of this approximation on classification tasks was not analyzed.

*Prototype-based Classification.* The idea of using a set the prototypes and a similarity function to implement a text classifier has a long tradition in information retrieval (Joachims, 2002; Sebastiani, 2002). It is also a popular method to reduce the complexity of nearest neighbor approaches (Garcia et al., 2012). In particular, centroid based text classification (CC), where class dependent centroids are used as prototypes, has been recently revisited by several authors because of the computational efficiency that can be achieved in large scale applications. Tan (2008) reports significant improvements on naive Bayes and KNN methods using an adaptive centroid classifier for text categorization tasks. An online extension of this procedure has been presented in (Tan et al., 2011) and applied to large train/test problems where the method slightly outperforms SVMs. A similar technique was presented in (Borodin et al., 2013) for text classification in stationary and non-stationary environments. Wang et al. (2013a) propose to filter out instances far from the boundary to enhance the predictive power of CC and Pang and Jiang (2013) integrate it with a clustering algorithm to obtain a lightweight approximation of nearest neighbors. Unfortunately, all these contributions focus on single-label classification. Multi-label annotations are indeed filtered out for experimental purposes and thus it is not yet well known how to adapt CC to multi-label scenarios. In this paper, we explicitly devise a multi-label classification method to annotate documents, using centroids to learn a partition of the representation space where documents has been embedded using an online representation method.

## 6. EXPERIMENTS

This section presents experiments performed to evaluate the performance of the clashing system and other text classification methods, explicitly devised or adapted for data stream settings.

We start our analysis studying the ability of the online embedding procedure studied in this paper to effectively preserve the geometry of the word count space. In particular, we seek to determine how the performance of the clashing approach depends on the dimensionality of the low-dimensional representation space. Next, we perform experiments to compare the performance of system with other text classifiers. The performance measures used in all the experiments of this section are detailed below.

*6.1. Performance Measures*

Widespread metrics to assess performance text classification are precision and recall. *Precision* can be defined as the probability that a retrieved instance is relevant to a given query and *Recall* as the probability to retrieve a relevant instance (Joachims, 2002). Given a class label $\tau_j$ they be respectively estimated as follows

$$\hat{p}_j = \ n^j_{++}/(n^j_{++} + n^j_{+-}) \ , \ \hat{r}_j = \ n^j_{++}/(n^j_{++} + n^j_{-+}),$$

where $n^j_{+-}$ is the number of documents for which $\tau_j \in f(\mathbf{x}_i)$ but $\tau_j \notin T_i$ (false positives), $n^j_{++}$ is the number of documents for which $\tau_j \in f(\mathbf{x}_i)$ and actually $\tau_j \notin T_i$ (true positives), $n^j_{-+}$ is the number of documents for which $\tau_j \in T_i$ but $\tau_j \notin f(\mathbf{x}_i)$ (false negatives). Usually, a tradeoff between precision and recall is unavoidable (Joachims, 2002). For this reason, these scores are usually combined into a single performance measure, called the $F_1$ measure

$$F_1^j = 2p_j r_j/(p_j + r_j) = 2n_{++}^j/(2n_{++}^j + n_{+-}^j + n_{-+}^j). \qquad (21)$$

In multi label classification, the goal is to obtain a good performance among all the possible class labels, including those with less samples. Two different methods are typically used to assess such multi-label performance: macro- and micro- averaging. In macro-averaging, performance is measured by averaging the values of $F_1^j$ among the different labels, $F_1^{\text{macro}} = n_t^{-1} \sum_j F_1^j$. In micro-averaging in contrast, the different types of errors are first computed as a whole, and just then they are processed to compute a value $F_1^{\text{micro}} = 2n_{++}/(2n_{++}+n_{+-}+n_{-+})$, where $n_{+-} = \sum_j n_{+-}^j$, $n_{++} = \sum_j n_{++}^j$ and $n_{-+} = \sum_j n_{-+}^j$. These two methods may give quite different results. Micro-averaging tends to overwhelm labels with less samples among data. Macro-averaged measures are usually more difficult to optimize because they give equal importance to all the class labels, including those which are under-represented with respect other labels (Tsoumakas et al., 2010).

*6.2. Preservation of Metric Information*

In this experiment we study the preservation of the metric information on the Reuters RCV1 Volume I Corpus (RCV1). This an archive of over 800.000 newswire stories, collected and manually categorized by Reuters, extensively used to assess text mining algorithms (for details see Lewis et al., 2004). We use the Topics category set consisting of 103 tags which capture the major subjects of a story. For this corpus, we compute two $n \times n$ Gram matrices $\mathbf{M}^{\mathbf{x}}$ and $\mathbf{M}^{\phi}$ composed of the inner products between documents in the original vector space $\mathcal{X}$ and the low dimensional representation space $\phi(\mathcal{X})$ respectively; that is, $\mathbf{M}_{ij}^{\mathbf{x}} = \langle \mathbf{x}_i, \mathbf{x}_j \rangle$ and $\mathbf{M}_{ij}^{\phi} = \langle \phi(\mathbf{x}_i), \phi(\mathbf{x}_j) \rangle$. As we have discussed before, a document classification system based on TF-IDF vector space representation is typically more accurate than a system based only on TF. However, direct embedding of TF-IDF requires precomputing a entire vocabulary for the corpus beforehand, thus preventing a fully online operation of the proposed system. This experiment is also aimed to determine if the IDF correction scheme we have described in previous sections allow us to get correlation back.

Figure 1 shows the linear correlation coefficient $\rho$ as a function of the number of the number of dimensions (in logarithmic scale) obtained by sampling $2 \times 10^6$ pairs of points from the $\mathbf{M}^{\mathbf{x}}$ and $\mathbf{M}^{\phi}$, for the different vector space representations. As it may be expected, the linear correlation between TF and the hashed TF representation (red, circled, solid curve) converges to 1 as the number of dimensions increases. After $m = 2^{14}$ dimensions, the linear correlation is practically optimal, and even $m = 2^{12} = 4096$ dimensions provides a quite satisfactory level of approximation ($\rho = 0.9240$). The correlation between TF-IDF and hashed TF representations (green, dashed, circled curve) in contrast, is convergent to approximately 0.95, demonstrating that IDF indeed changes the geometry of the TF vector space. However, after applying the IDF correction scheme to the hashed TF representations (orange, diamond, dotted curve) the linear correlation is convergent to 1 again, as desired, thus confirming that the TF-IDF vector space can be approximated by hashing without storing a vocabulary.

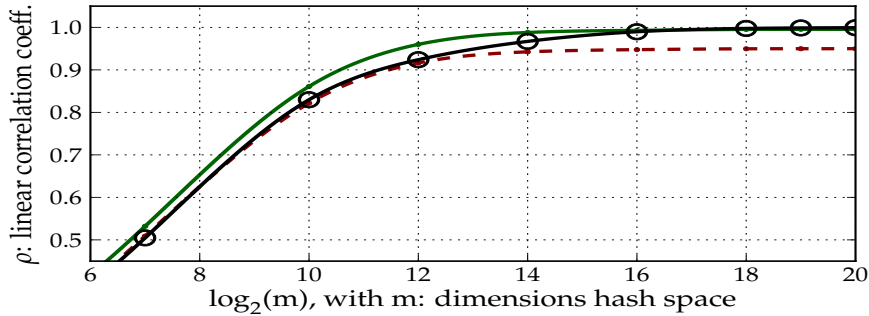

Figure 1: Linear correlation coefficient between the inner products in the original vector space and the hashed space. For the solid curve, the original space corresponds to a TF representation and the reduced space is $\phi(\mathcal{X}_{\text{tf}})$. For the dashed curve, the vector space $\mathcal{X}_{\text{tf}\times\text{idf}}$ is built by using TF-IDF, but the hashed space is still $\phi(\mathcal{X}_{\text{tf}})$. For the circled solid curve, the hashed space is our approximation to $\phi(\mathcal{X}_{\text{tf}\times\text{idf}})$.

*6.3. Effect of Dimensionality on the Classification Performance*

Continuing with the previous experiment, we study the predictive power of the clashing models as a function of the dimensionality of the hashed space. With this aim, we use all the documents dated in odd days as an independent test set and the rest are used as training instances. The training instances are processed in chronological order, following the data stream setting introduced in section 2.2, that is, each document is classified and used as a training instance only once. Figure 2 shows the Macro a Micro F1 measures on the test set after processing the training documents and considering all the 103 tags in the Reuters Corpus (Topics collection).

After $m = 2^{10}$ dimensions, we observe a fast convergence to a stable performance. Note that this is coherent with the results obtained in the previous section in which we observed a fast convergence to linear correlation $\rho = 1$ after $m = 2^{12}$ or $m = 2^{14}$ dimensions. Indeed, these results suggest that in order to obtain a competitive performance, the system does not need to preserve *all* the geometric information of the original space and it is instead robust to slight distortions in the inner products computed in the compressed space.

*6.4. Batch Sanity Checks*

We present a classic train-test experiment aimed to compare the predictive performance of the models analyzed in this paper with some results available in (Lewis et al., 2004), in which the authors thoroughly study the Reuters corpus. Table 1 shows accuracy, macro and micro averaged F measure, precision and recall, computed on the test set, for different classifiers, after processing the full labelled data stream. For kNN, we set $k = 1$ and predicted all the labels contained in the nearest neighbor of a test document. Binary models (perceptron and SVM) were trained using a binary relevance approach (Sebastiani, 2002; Lewis et al., 2004). The perceptron was trained sequentially on the dataset with a learning rate of 0.1. SVMs were trained using the LIBLINEAR library for large-scale classification (Fan et al., 2008), using a classic L1-loss, L2-regularized formulation with default parameters. Following Lewis et al. (2004), a first SVM (SVM-Lewis) was trained using the first 22.000 training examples in chronological order. SVM-Full was trained using the full training stream (400K examples).

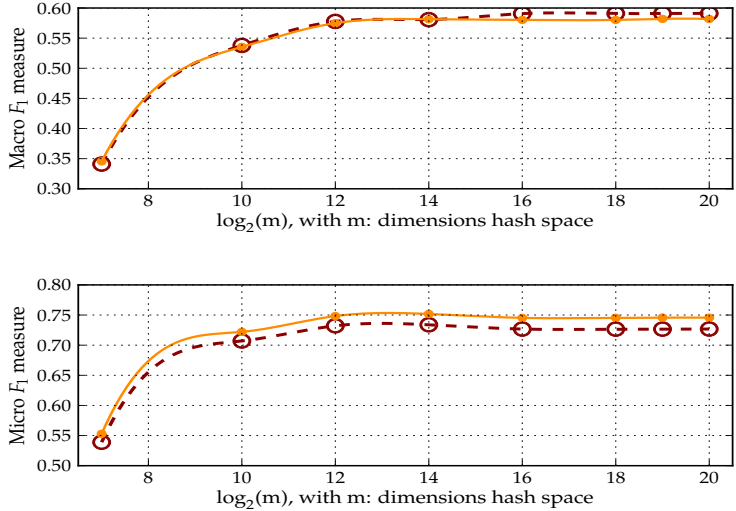

Figure 2: Macro a Micro F1 measures on the test set as a function of the number of dimensions in the hash space. The learning mode-1 is shown in dotted curves (circles at data points) and the learning mode-2 using solid curves (points at data points).

We conclude that the performance of kNN and SVM models is similar or slightly better than those reported in (Lewis et al., 2004). The SVM-Full is clearly the best classifier in this setting, followed by kNN. However, these models require significantly more computational resources than a clashing system. In particular, kNN has a training complexity of $\mathcal{O}(Dn^2)$, where $D$ is the dimensionality of the representation space and $n$ is the size of the training set. Similarly, SVMs exhibit a training time between linear and quadratic in $n$ even using modern solvers.

Using few computational resources, the clashing models achieve macro averaged scores competitive with those of kNN. The Clasher-M2 shows a better macro averaged recall than kNN, as expected from its design. In terms of macro F measure, both Clashers improve on the SVM trained with the amount of data used in (Lewis et al., 2004) to make training tractable. The online perceptron achieves a lower macro F measure due to a lower macro averaged precision. This scenario is similar in terms on micro averaged measures, except for the perceptron and SVM-Lewis which become significantly more competitive in terms of precision and F measure. This improvement with respect to macro averaged measures suggests that these models are better for learning classes observed more frequently among documents, underwhelming classes with less samples.

|            | **Macro** |        |        |        | **Micro** |        |        |
|------------|-----------|--------|--------|--------|-----------|--------|--------|
| **Classifier** | **F**     | **P**  | **R**  | **Acc** | **F**    | **P**  | **R**  |
| Clasher-M1 | 0.5774    | 0.6145 | 0.5972 | 0.9847 | 0.7322    | 0.8039 | 0.6720 |
| Clasher-M2 | 0.5740    | 0.6012 | 0.6319 | 0.9855 | 0.7476    | 0.8159 | 0.6899 |
| kNN        | 0.6114    | 0.6192 | 0.6067 | 0.9860 | 0.7767    | 0.7756 | 0.7778 |
| Perceptron | 0.5495    | 0.5305 | 0.5860 | 0.9838 | 0.7462    | 0.7313 | 0.7617 |
| SVM-Lewis  | 0.5394    | 0.6942 | 0.4734 | 0.9872 | 0.7812    | 0.8355 | 0.7336 |
| SVM-Full   | 0.65173   | 0.7244 | 0.6049 | 0.9887 | 0.8130    | 0.8427 | 0.7854 |

Table 1: Performance measures on the test set: F measure (F), Precision (P), Recall (R) and Accuracy (Acc).

*6.5. Data Stream Experiments*

In this section, we analyze the cumulated performance of the clashing models and alternative classification methods in the data stream setting described in section 2.2. At each round $t$, the system is given with a new document $d_t \in \mathcal{D}$ and it is asked to predict the set of labels $\hat{T}_t \subset \mathcal{T}$ that should be assigned to it. After providing a prediction $\hat{T}_t = f_{t-1}(d_t)$, the system is feed with the set of correct labels $T_t$ and this information is used to update the model. The *cumulative performance* achieved by the models is monitored by updating counters for false positives, true positives, false negatives and true negatives, which are then used to compute precision, recall, and $F_1$ measures at any desired time. Since we are interested in a multi-label evaluation of the classifiers, we compute micro and macro averages among all the 103 tags of the Reuters corpus. That is, for a micro average we use global errors counts. For obtaining a macro average, we first compute the measure of performance for each possible tag and then average the results.

In addition to the Reuters data, we present here scalability experiments performed in the full New York times annotated corpus (Sandhaus, 2008). This is a collection of 1.855.658 articles obtained from the historical archive of The New York Times, covering a period of more than twenty years, between January 1987 and June 2007. According to Crammer et al. (2009), this is possibly the largest collection of publicly released annotated news text, and therefore an ideal benchmark to test large-scale text analysis tools. In this paper we select the *General Online Descriptors* family of labels to illustrate the methods (1622 tags).

Figure 3 displays the cumulative performance measures for the different models as they process the Reuters corpus and Figure 5 shows the corresponding results using the New York Times dataset. We compare the average running time (secs.) required by the models to process one document, including testing and training operations. This time was measured every 5000 rounds. The second and third panels show the macro-averaged and micro-averaged $F_1 measure$ respectively. These scores were computed every 500 rounds for the first 20.000 rounds and every 5000 rounds after that. Additionally we show in Figure 4 a comparison of the clashing system, operating with the simplest of the two learning criteria investigated (Mode-1), against the online Hoeffding tree models proposed in (Read et al., 2012) to deal with multi-label data streams.

*Clashing versus kNN and SVMs.* . For this experiment, a SVM was periodically re-trained using batches of increasing size extracted from the stream. An initial model was computed using the first 4000 observations and re-computed every 2000 examples during the first 20.000 test/train rounds. Then, we increase the training period to 20.000 examples. As for the previous experiment, we employ a L1-loss, L2-regularized formulation, solved using the large-scale coordinate ascent method provided in the LIBLINEAR library (Fan et al., 2008), using default parameters. kNN was used by setting $k = 1$ and storing all the labelled documents in memory as they arrive to the system.

In Figure 3 (Reuters data), we observe that all the models quickly converge to a micro averaged $F_1$ measure close to the test performance observed in the previous experiment. The SVM is significantly superior in terms of this score, followed by the kNN approach. However, in terms of macro averaged $F_1$ measure (second panel of Figure 3 ), the picture is very different. The clashing models provide superior predictive performance uniformly in the first half of the stream. kNN and the SVM need to observe significantly more observations in order to converge to performances comparable to those observed in the batch setting. This result confirms that our models are more effective for modeling classes underrepresented in terms of samples. By comparing the macro averaged and micro averaged $F_1$ measures, we conclude that kNN and the SVM overwhelm more easily the popular labels at expenses of tags with less samples. The different implementations of the learning rule for the clashing system do not show significant differences.

From Figure 5 (New York Times data), we observe that the clashing models obtain significantly better performance in terms of macro averaged $F_1$ measure than the SVM. The SVM performance indeed starts to decrease at $n \approx 200K$ documents and only after $n \approx 800K$ it starts to slowly increase again. Studying the learning curves corresponding to micro averaged $F_1$ measure, we note that the performance of all the models start to decrease around $n \approx 200K$ documents. This trend starts to be slowly reversed around $n \approx 800K$ documents for the SVM. We explain this result by hypothesizing a concept drift between $n \approx 200K$ and $n \approx 800K$ for the classes of the problem which are dominant in terms of number of samples. We confirm this hypothesis by studying the performance of the clashing models on a random permutation of the data stream. From Figure 5, we observe, as expected, that the performance of the model is practically constant on the randomly shuffled data.

Since the micro averaged $F_1$ measure is highly biased towards the performance obtained in the prediction of over represented labels, all the models suffer the effect of the concept drift if we measure performance according to this score. After the drift stops, the SVM is able to more quickly increase its micro averaged score because it is more easily affected by the classes with more samples. The clashing models do not recover easily after the drift stops but their performance decreases more slowly after it starts, in such a way that they exhibit a performance comparable to the SVM at the end. Since the clashing models are not biased towards the most popular labels, they obtain monotonically increasing macro averaged $F_1$ measures, in contrast to the SVM which suffer the drift also in terms of the micro averaged $F_1$ score.

The first panel of Figure 3 illustrates that the running time complexity of the clashing models is significantly better than the time required by kNN and the SVM to process examples. kNN scales linearly with the data stream size because it stores each new document presented to the system. Even using a modern solver, running time of the the SVM seems to scale linearly or super-linearly in the number of processed documents. The same conclusion is obtained from Figure 5: processing times for the SVM seems to scale linearly in the stream size. Our methods in contrast guarantee constant processing time in both cases.

*Clashing versus Online Perceptron.* . This comparison is interesting because, as the proposed methods, the perceptron is a truly online method.

In Figure 3 (Reuters data), we note that the micro averaged $F_1$ measure achieved by the perceptron is similar to that of the clashing techniques with an advantage at the end. The perceptron is slightly worse than the proposed methods on the first part of the stream. However, the predictive power of the clashers increase significantly faster than the perceptron's learning curve when we analyze the macro averaged $F_1$ measure. Our models keep a significant advantage most of the time. This result suggests again that the clashing approach is better to predict tags which are underrepresented in terms of samples as compared to more frequently observed tags. Underwhelming classes with a high number of negatives with respect to positives (i.e. documents containing the label) is a common problem in multi-label applications, in such a way that obtaining high macro averaged scores is usually hard in practice.

In Figure 5, we observe that the trend of the perceptron's learning curve on the New York Times data is similar to form the SVM's learning curve. However, most of the time, the perceptron achieves an $F_1$ measure lower than the SVM's and thus lower than the clashing models'. Our previous conclusions about a likely concept drift between $n \approx 200K$ and $n \approx 800K$ hold as well. The perceptron is more readily affected by the drift and it's able to more quickly recover from it. We explain this result by the tendency of the perceptron to overwhelm classes with a high number of samples, i.e., with a large number of positive examples.

From Figure 3 (Reuters data), we note that, at the beginning, the average processing time for this technique is higher and more variable because the probability of incurring a classification loss is higher. This suggests that in applications with more complex, e.g. drifting, concepts, the efficiency of this technique may decrease. The running time of the clashing models is instead always constant, independently of the current performance. In Figure 5, we observe that the processing times of the the clashing models on the New York Times data are significantly better than the perceptron's times. This behavior can be explained by the low performance of the perceptron in terms of averaged $F_1$ measures which translate into a high number of mistakes triggering updates, and the large number of sub-models the perceptron needs to handle in this dataset (1622 tags).

*Clashing versus Online Hoeffding Trees.* . In figure 4, we show the results achieved on the Reuters data stream of 4 techniques investigated in (Read et al., 2012) to deal with multi-label data streams. Online Hoeffding trees with Majority Label classifiers at the leaves are denoted as HT-ML. Online Hoeffding trees with Pruned Label Sets at the leaves are denoted as HT-PS. As stressed by Read et al. (2012) these methods may be significantly enhanced by using them in ensemble methods like (Bifet et al., 2009) and (Oza, 2005) designed for data streams. Here we show the results obtained by using the method in (Bifet et al., 2009), but the results are similar to those obtained by using (Bifet et al., 2009). Ensemble versions of HT-ML and HT-PS are denoted EA-HT-ML and EA-HT-PS. We used the implementation provided by the authors in the last version of the MOA software (Bifet et al., 2010). Parameters were set as default, e.g., we used adaptive thresholding and $M = 10$ base learners. Results show that the clashing system is significantly more accurate than all these techniques both in terms of Macro F measure and Micro F measure. Running times are low for all the techniques with a slight advantage for the method investigated in this paper.

*6.6. Partially Labelled Streams*

From the results presented so far, we hypothesize that the clashing system can be advantageous learning from a partially labelled stream because of its ability to reach a competitive performance more quickly. To confirm this hypothesis we conducted an experiment on the full Reuters dataset, where the classifiers are trained using different fractions ($p$) of labelled data. We present the 800.000 examples corresponding to this dataset in a series of rounds $t = 1, 2, \ldots$. Every document is used as testing instance to compute the cumulative performance measures. However, after testing, a document is used as a training instance with probability $p$.

Figure 6 displays the results obtained by setting, $p = 0.25$, $p = 0.125$, $p = 0.0625$ and $p = 0.03125$. These results confirm that the clashing approach is able to learn faster than the other methods from a few labelled instances. This advantage is more clear when we focus on macro averaged F measure.

## 7. CONCLUSIONS

In this paper, we presented a simple and efficient method to classify large data streams of documents, referred to as clashing.

*Contributions.* The novelty of the proposed approach is in addressing both text representation and document classification in an online fashion, using limited computational resources. This is a challenging setting because: (1) the computation of standard text representation requires to known in advance the documents and words to be processed, (2) an exact incremental text representation based on words requires unbounded resources, (3) usually text annotations are non-exclusive but most multi-label classifiers have been studied in batch scenarios.

We solved the problem of text representation by devising a novel method to approximate the geometry of the full TF-IDF space using only online computations. This approach combines recent ideas of data stream sketching and feature hashing. Up to our knowledge, this is the first paper showing that the online computation of both TF and IDF, without corpus-wise periodical computations, is possible and accurate. To efficiently predict multi-label classifications, we devised a method to organize the feature space into a set of regions where documents with similar low dimensional representations collide by way of a winner-takes-all mapping process. Then, we implemented a conditional naive bayesian approach where labels are assumed to be independent given the clashing region. This method allows to keep the high efficiency of the text representation engine. Up to our knowledge, this is the first extension of centroid based classification to multi-label data and data stream text classification. We showed that this method can be simple, scalable and accurate.

The clashing system showed constant processing time independently of the data stream size and the number of possible attributes (words). We performed experiments using the Reuters RCV1 and New York Times data. They suggest that learning from data streams using this system is efficient and more robust to unbalanced classes than methods with comparable running times, yielding better macro averaged predictive scores. Finally, the system arrives faster to give forth competitive predictions even when it has observed few labelled instances. These properties can be useful in settings where the classifier needs to learn from partially labelled streams or in drifting scenarios, where components of the model need to quickly learn from new observations to fit the new data configuration.

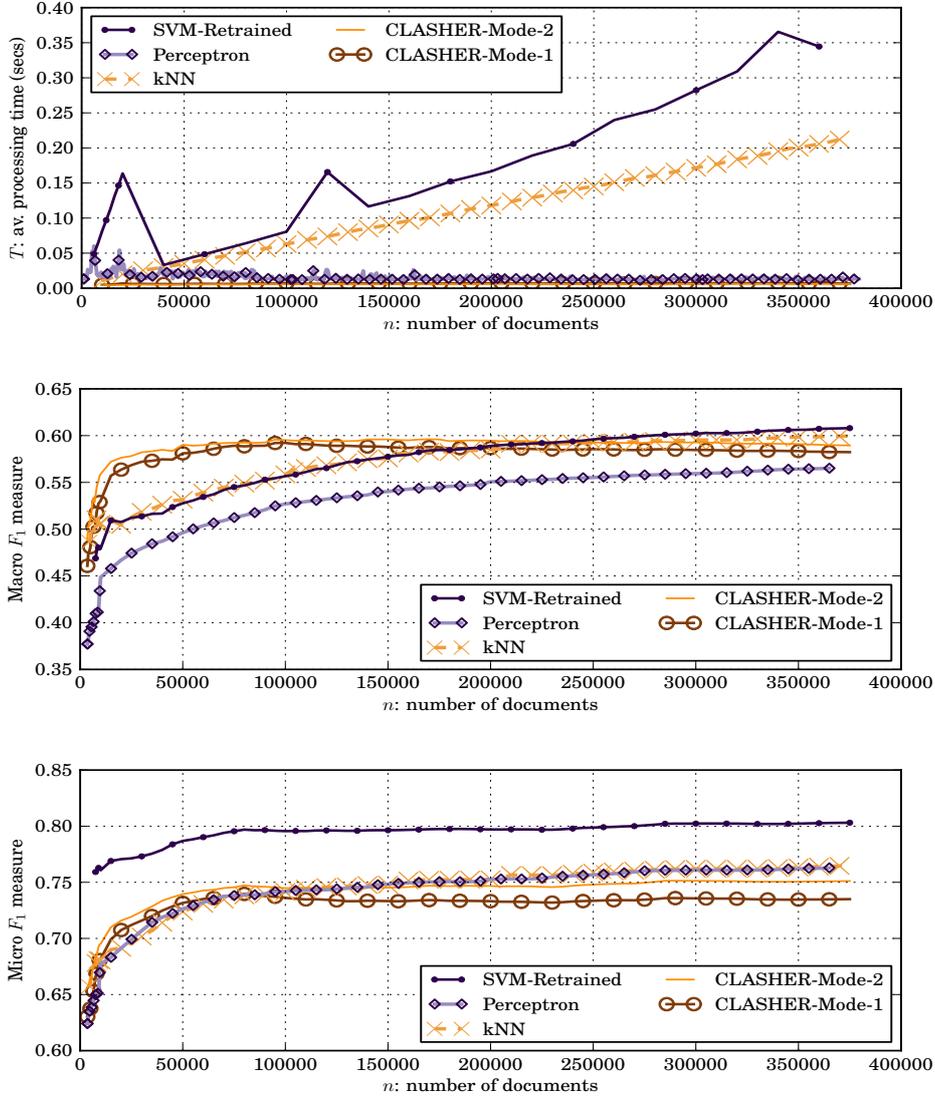

Figure 3: Results obtained in the Reuters Corpus. (From top to bottom) First Panel: Average time required to process a training document as of time. Second Panel: Macro averaged F measure among all the labels. Third Panel: Micro averaged F measure.

*Practical Implications and Future Work.* This work was motivated by a real world problem in which we need to annotate and analyze large streams of web content. The algorithm we have developed will be incorporated into our pipeline devoted to Computational

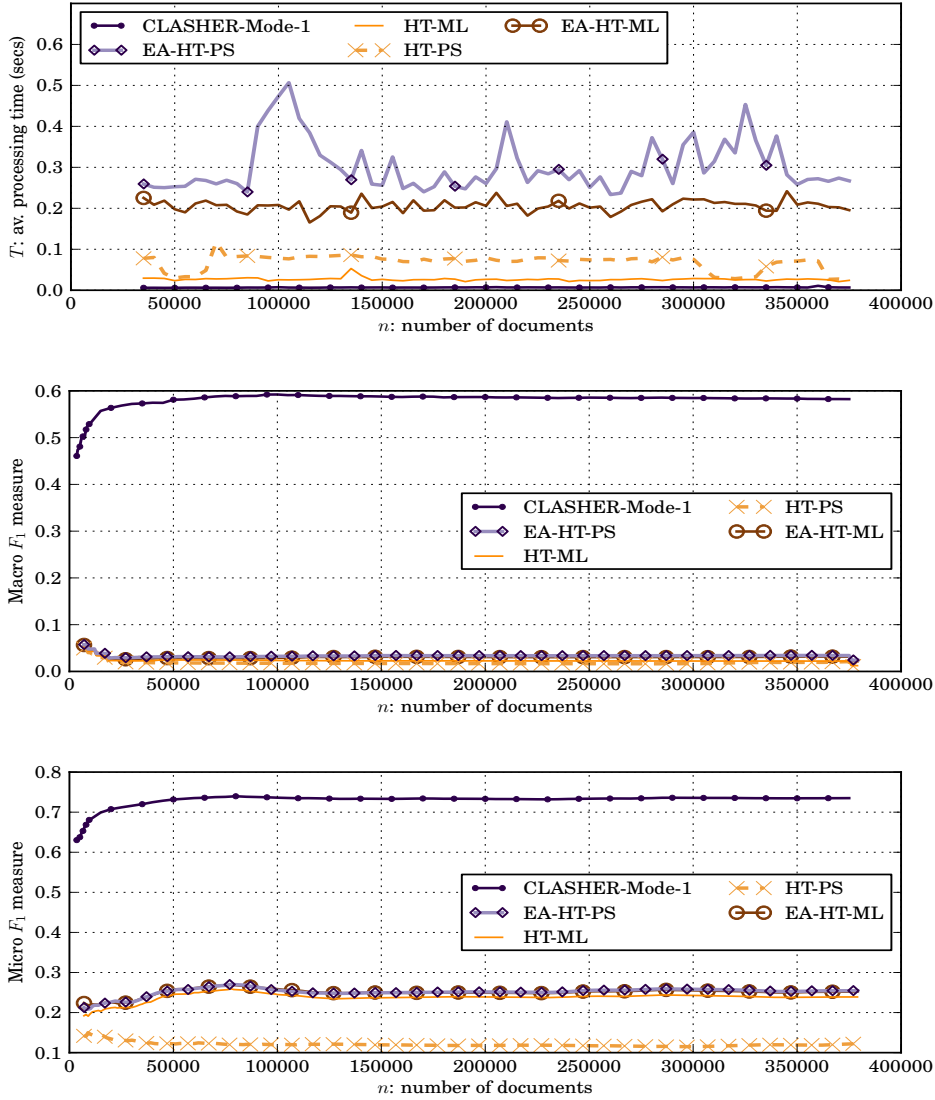

Figure 4: Comparison with Different Variants of the Online Multi-label Hoeffding Trees introduced in Read et al. (2012). (From top to bottom) First Panel: Average time required to process a training document as of time. Second Panel: Macro averaged F measure among all the labels. Third Panel: Micro averaged F measure.

Social Sciences, NOAM (Flaounas et al., 2011). The insistence on only making use of bounded resources is a consequence of the size of the streams: both time and memory need to be kept under control. This problem comes up often in practice. Fast flowing streams of text are generated by online news, social media and endless other applications, and the need to automatically annotate and adaptively sort them into sub-streams is a crucial one.

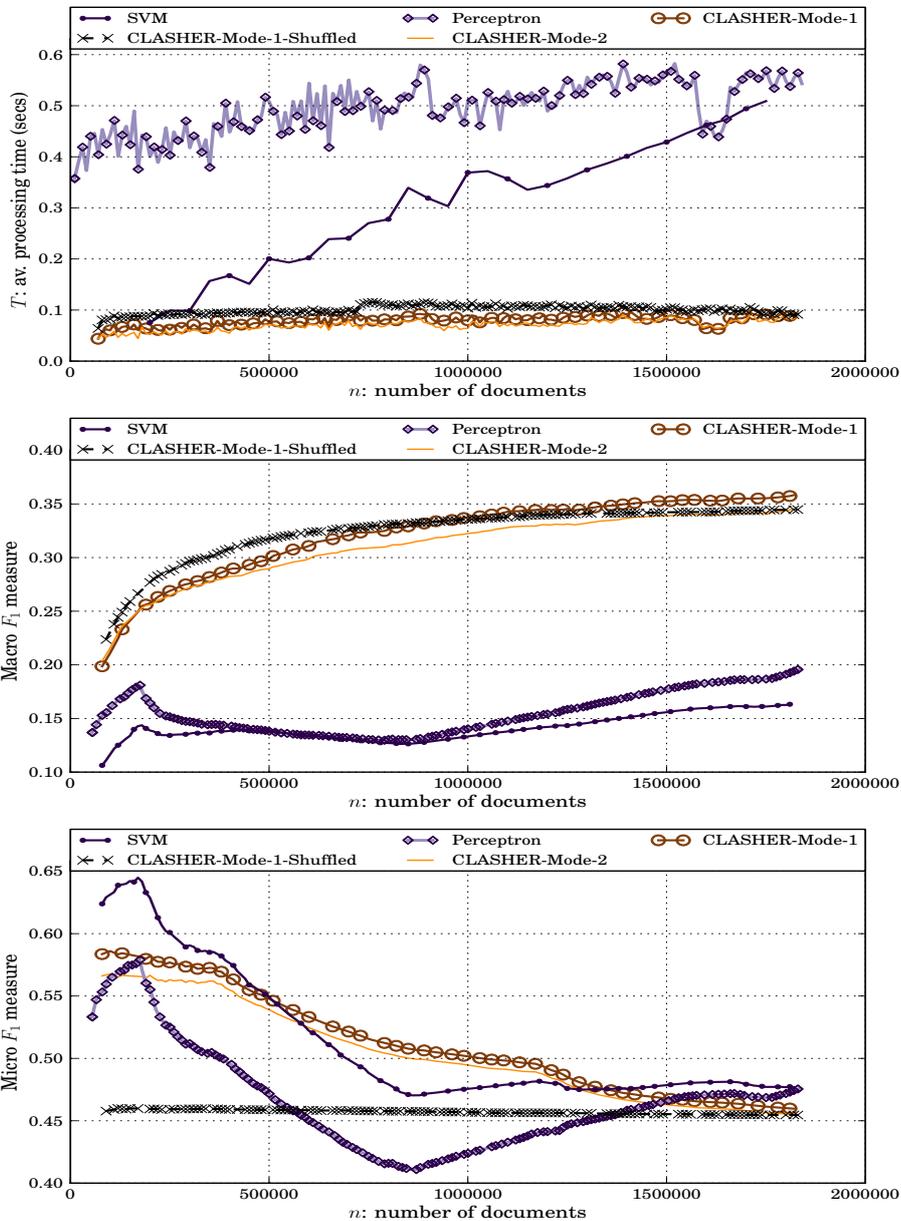

Figure 5: Results obtained in the New York Times Corpus. (From top to bottom) First Panel: Average time required to process a training document as of time. Second Panel: Macro averaged F measure among all the labels. Third Panel: Micro averaged F measure.

In the future, we plan to extend this work in the following directions. (1) To incorporate in the system the ability to explicitly detect and manage concept drift. We believe that the system is particularly suitable to support this feature by incorporating adaptive forgetting factors in the engines used to approximate TF-IDF and to obtain multi-label annotations. (2) To use our online text representation system in the solution of unsupervised text mining problems. (3) To extend the online embedding procedure to approximate other weighting schemas. In particular class-based term weighting for text classification has been focus of increasing interest in the last years (Lertnattee and Theeramunkong, 2004; Lan et al., 2009; Guan et al., 2009; Luo et al., 2011; Ren and Sohrab, 2013; Wang and Zhang, 2013). Unfortunately, latest contributions typically adopt a batch setting for text representation. Our preliminary analysis suggests that some approaches, e.g. (Ren and Sohrab, 2013), may be efficiently and accurately approximated using data sketches without significantly increasing storage and processing time.

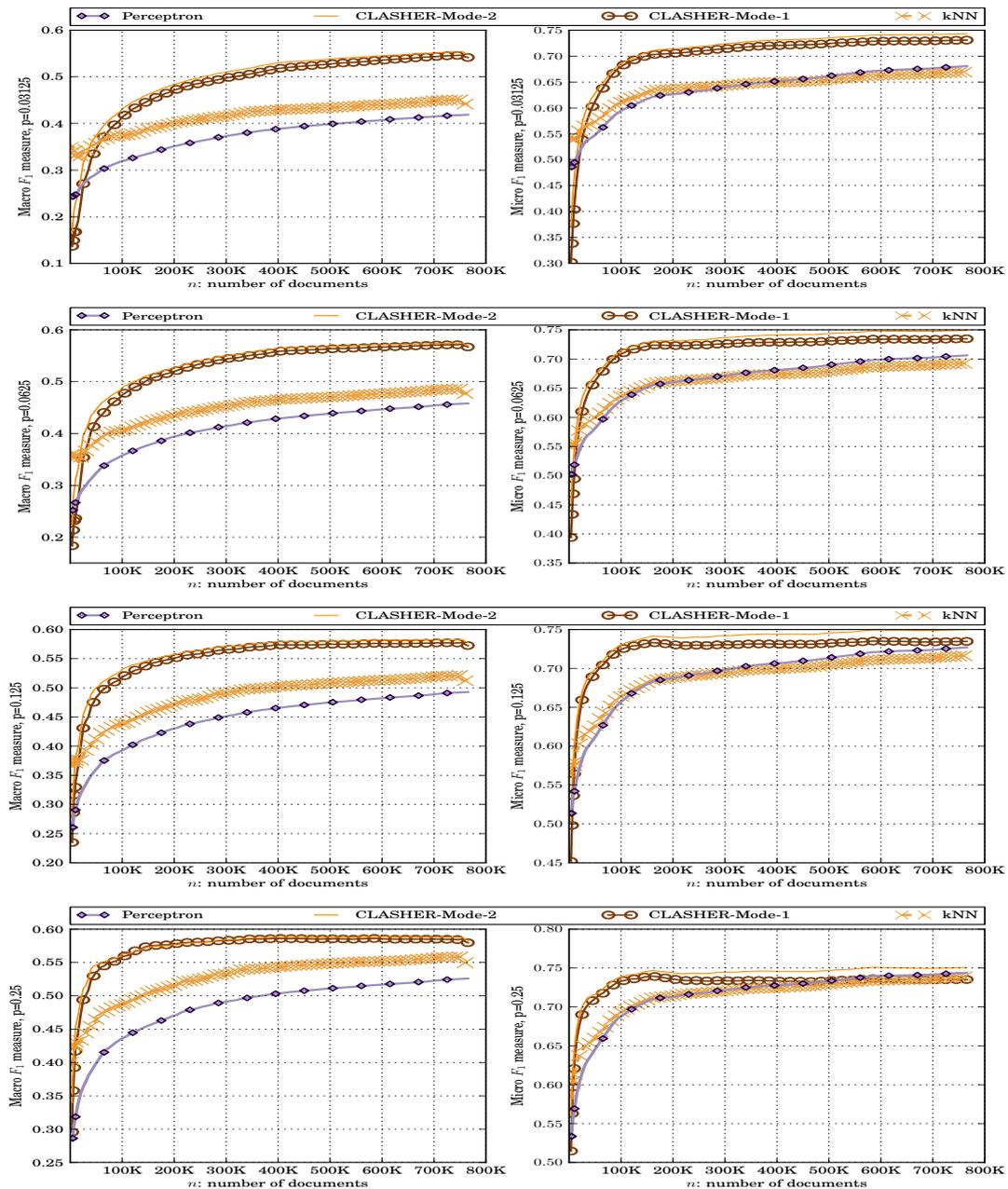

Figure 6: Predictive performance under different fractions ($p$) of labelled data on the full data stream. (From top to bottom) $p = 0.03125$, $p = 0.0625$, $p = 0.125$ and $p = 0.25$. The figures at the right (left) correspond to macro-averaged (micro-averaged) F measures.

## Acknowledgments

Nello Cristianini and Ilias Flaounas were supported by EU project COMPLACS. Ricardo Ñanculef was partially funded by the National Commission for Scientific and Technological Research of Chile, Grant Fondecyt 11130122.

**APPENDIX (PROOFS)**

*Proof.* (Lemma 1 ). We have shown in 3.1, that the array $C_k$ computed by Alg. 3 corresponds to an implementation, with $L=1$, of the Cormode and Muthukrishna sketch, described in 2.5, to approximate the document frequencies ($f(w_i)$) of the words observed till round $t$. Therefore, for any word $w_i$ such that $h(w_i) = k$, we have from Theorem 2, $f(w_i) \leq C_k \leq f(w_i) + \epsilon \|\mathbf{f}\|$, with probability $(1 - e^{-1})$, provided $m \geq \lceil \exp(1)/\epsilon \rceil$. Now, since $\log(\cdot)$ is a monotonically increasing function.

$$f(w_i) \leq C_k \leq f(w_i) + \epsilon \|\mathbf{f}\|$$
$$\Rightarrow \log(f(w_i)) \leq \log(C_k) \leq \log(f(w_i)) + \log\left(\frac{f(w_i) + \epsilon \|\mathbf{f}\|}{f(w_i)}\right)$$
$$\Rightarrow \log(f(w_i)) \leq \log(C_k) \leq \log(f(w_i)) + \log\left(1 + \frac{\epsilon \|\mathbf{f}\|}{f(w_i)}\right). \tag{22}$$

The left hand side implies

$$\log\left(\frac{n}{f(w_i)}\right) = \log(n) - \log f(w_i) \geq \log(n) - \log(C_k) = \log\left(\frac{n}{C_k}\right). \tag{23}$$

Thus $\text{IDF}(w_i) \geq \overline{\text{IDF}}(w_i)$. Similarly, the left hand side of (22) implies (we assume $f(w_i) > 1$ for simplicity)

$$\log\left(\frac{n}{C_k}\right) \geq \log\left(\frac{n}{f(w_i)}\right) - \log(1 + \epsilon \|\mathbf{f}\|). \tag{24}$$

Thus $\overline{\text{IDF}}(w_i) \geq \text{IDF}(w_i) - \bar{\epsilon} \|\mathbf{f}\|$, with $\bar{\epsilon} := \log(1 + \epsilon \|\mathbf{f}\|) / \|\mathbf{f}\|$. Solving for $\epsilon$ yields $\epsilon = (\exp(\bar{\epsilon}\|\mathbf{f}\|) - 1)/\|\mathbf{f}\|$. Using the inequality $\exp(x) \geq x + 1$ which holds for any $x < 1$, we have $\epsilon = (\exp(\bar{\epsilon}\|\mathbf{f}\|) - 1)/\|\mathbf{f}\| \geq ((\bar{\epsilon}\|\mathbf{f}\|) + 1 - 1)/\|\mathbf{f}\| = \bar{\epsilon}$. Thus $\epsilon \geq \bar{\epsilon} > 0$. Therefore, $\overline{\text{IDF}}(w_i) \geq \text{IDF}(w_i) - \bar{\epsilon} \|\mathbf{f}\|$ implies $\overline{\text{IDF}}(w_i) \geq \text{IDF}(w_i) - \epsilon \|\mathbf{f}\|$. □

*Proof.* (Lemma 2). Let $X = \{\mathbf{x}_1, \ldots, \mathbf{x}_n\}$ the representation of the documents seen so far in the word-count space $\mathcal{X}$. Let $X_i := \{\mathbf{x}_k \in X : \tau_i \in T_k\}$ and $\mathbf{X}_i$ the matrix with columns corresponding to the elements of $X_i$. Let $\mathbf{S}^\phi$ and $\mathbf{S}^\mathbf{x}$ denote the clashing systems operated on $\phi(\mathcal{D})$ and $\mathcal{X}$ respectively. Let $\bar{\mathbf{p}}_1, \ldots, \bar{\mathbf{p}}_{n_t}$, with $\bar{\mathbf{p}}_i = \mathbf{X}_i \boldsymbol{\alpha}$, represent the prototypes of $\mathbf{S}^\mathbf{x}$ and $\bar{C}_{i,j} = |X_i \cap X_j|/|X_i|$ its probability estimates. Here, $\boldsymbol{\alpha}$ is the set of coefficients used by the system to linearly generate the prototypes. Note that the Mapper in $\mathbf{S}^\mathbf{x}$ computes $j* = \arg\min_i \|\mathbf{x} - \bar{\mathbf{p}}_i\|^2$ and the prediction rule given by Eqn. (13) still holds by using the new definition of $j*$. Note now that due to the linearity of the hashed feature map,

$$\mathbf{p}_i = \phi \mathbf{X}_i \boldsymbol{\alpha} = \phi(\mathbf{X}_i \boldsymbol{\alpha}) = \phi(\bar{\mathbf{p}}_i). \tag{25}$$

This allows us to use the bounds for preservation of norms in $\mathcal{X}$ given in the previous section. Suppose, $\mathbf{x}$ is assigned to the region $i*$ in the original data space. For any $\delta < 0.1$, let $\eta_j(\mathbf{x}) = \|\bar{\mathbf{p}}_j - \mathbf{x}\|^2 - \|\bar{\mathbf{p}}_{i*} - \mathbf{x}\|^2$, $\eta(\mathbf{x}) = \min_{j \neq i*} \eta_j(\mathbf{x})$ and $\epsilon_j = \frac{\eta(\mathbf{x})}{2\|\bar{\mathbf{p}}_j - \mathbf{x}\|^2}$. Consider the event $\mathcal{A}^\phi(\mathbf{x})$ composed of the $\phi(\mathbf{x})$ such that (simultaneously)

$$\|\mathbf{p}_{i*} - \phi(\mathbf{x})\|^2 = \|\phi(\bar{\mathbf{p}}_{i*}) - \phi(\mathbf{x})\|^2 \leq \|\bar{\mathbf{p}}_{i*} - \mathbf{x}\|^2 + \epsilon_{i*} \|\bar{\mathbf{p}}_{i*} - \mathbf{x}\|^2 \tag{26}$$
$$\|\mathbf{p}_j - \phi(\mathbf{x})\|^2 = \|\phi(\bar{\mathbf{p}}_j) - \phi(\mathbf{x})\|^2 \geq \|\bar{\mathbf{p}}_j - \mathbf{x}\|^2 - \epsilon_j \|\bar{\mathbf{p}}_j - \mathbf{x}\|^2, \forall j \neq i*.$$

For any $j \neq i*$ and for any $\mathbf{x} \in \mathcal{A}^\phi(\mathbf{x})$ we obtain, by definition of $\epsilon_j$, $\eta_j(\mathbf{x})$ and $\eta(\mathbf{x})$

$$\|\mathbf{p}_{i*} - \phi(\mathbf{x})\|^2 \leq \|\bar{\mathbf{p}}_{i*} - \mathbf{x}\|^2 + \eta(\mathbf{x})/2 \leq \|\bar{\mathbf{p}}_{i*} - \mathbf{x}\|^2 + \eta_j(\mathbf{x})/2 \tag{27}$$
$$\|\mathbf{p}_j - \phi(\mathbf{x})\|^2 \geq \|\bar{\mathbf{p}}_j - \mathbf{x}\|^2 - \eta(\mathbf{x})/2 \geq \|\bar{\mathbf{p}}_j - \mathbf{x}\|^2 - \eta_j(\mathbf{x})/2,$$

and thus,

$$\|\mathbf{p}_{i*} - \phi(\mathbf{x})\|^2 \leq (\|\bar{\mathbf{p}}_{i*} - \mathbf{x}\|^2 + \|\bar{\mathbf{p}}_j - \mathbf{x}\|^2)/2 \tag{28}$$
$$\|\mathbf{p}_j - \phi(\mathbf{x})\|^2 \geq (\|\bar{\mathbf{p}}_{i*} - \mathbf{x}\|^2 + \|\bar{\mathbf{p}}_j - \mathbf{x}\|^2)/2,$$

which finally leads $\|\mathbf{p}_{i*} - \phi(\mathbf{x})\|^2 \leq \|\mathbf{p}_j - \phi(\mathbf{x})\|^2$, that is if $\mathbf{x} \in \mathcal{A}^\phi(\mathbf{x})$, the hashed decision for $\mathbf{x}$ is identical to the decision in the original data space. We just need to bound $\Pr(\mathcal{A}^\phi(\mathbf{x}))$. For any $\delta < 0.1$, set $\epsilon = \min_i(\epsilon_i)$, $m = 12\epsilon^{-2}\log(1/\delta)$ and $c = 16\epsilon^{-1}\log(1/\delta)\log^2(m/\delta)$. Note that this setting of parameters makes the conditions of Theorem 1 hold for any tuple $(\delta, \epsilon_i, m, c)$. Then, a simple union bound ensures that $\Pr(\bar{\mathcal{A}}^\phi(\mathbf{x})) \leq 3\delta n_t$. Rephrasing, the probability that the embedding keeps the decision made in the original data space is at least $1 - 3\delta n_t$. To conclude, note that to have a global confidence $\tilde{\delta}$ it is enough to choose $\delta = \tilde{\delta}/n_t$ and that

$$\bar{\eta}(\mathbf{x}) = \frac{\eta(\mathbf{x})}{\max_j \|\bar{\mathbf{p}}_j - \mathbf{x}\|^2} = 2\epsilon. \tag{29}$$

$\square$

*Proof.* Proposition 1) Let $\mathbf{x}$ be the exact TF-IDF representation of a document $d$ and $\mathbf{z}$ the its exact TF representation. If **IDF** denotes the vector of exact IDF weights corresponding to the coordinates of $\mathbf{x}$, we have $\mathbf{x} = \mathbf{IDF} \odot \mathbf{z}$, where $\odot$ denotes the Hadamard product (component by component). Alg. 3 computes an approximation $\bar{\mathbf{z}}$ of the TF representation $\mathbf{z}$, an approximation $\overline{\mathbf{IDF}}$ of $\mathbf{IDF}$ and finally approximates the TF-IDF representation as $\bar{\mathbf{x}} := \phi(d) = \overline{\mathbf{IDF}} \odot \bar{\mathbf{z}}$. Note that $\bar{\mathbf{z}}$ is just the hashed representation of $\mathbf{z}$ (identical to the embedding computed by Alg.2). In order to bound $|\|\mathbf{x}\|_2^2 - \|\bar{\mathbf{x}}\|_2^2|$, we can write,

$$\|\mathbf{x}\|_2^2 - \|\bar{\mathbf{x}}\|_2^2 = \left(\|\mathbf{x}\|_2^2 - \|\phi(\mathbf{IDF} \odot \mathbf{z})\|_2^2\right) + \left(\|\phi(\mathbf{IDF} \odot \mathbf{z})\|_2^2 - \|\bar{\mathbf{x}}\|_2^2\right), \qquad (30)$$

where $\phi(\mathbf{IDF} \odot \mathbf{z})$ is the hashed approximation of the true TF-IDF representation computed using Eqn.(4). Since $\mathbf{x} = \mathbf{IDF} \odot \mathbf{z}$, the first term is

$$\left(\|\mathbf{x}\|_2^2 - \|\phi(\mathbf{IDF} \odot \mathbf{z})\|_2^2\right) = \left(\|\mathbf{IDF} \odot \mathbf{z}\|_2^2 - \|\phi(\mathbf{IDF} \odot \mathbf{z})\|_2^2\right). \qquad (31)$$

Using Theorem 1, we have with probability $1 - 3\delta$

$$\left|\left(\|\mathbf{x}\|_2^2 - \|\phi(\mathbf{IDF} \odot \mathbf{z})\|_2^2\right)\right| \leq \epsilon \|\mathbf{x}\|_2^2. \qquad (32)$$

The second term is

$$\left(\|\phi(\mathbf{IDF} \odot \mathbf{z})\|_2^2 - \|\bar{\mathbf{x}}\|_2^2\right) = \left(\|\phi(\mathbf{IDF} \odot \mathbf{z})\|_2^2 - \|\overline{\mathbf{IDF}} \odot \phi(\mathbf{z})\|_2^2\right). \qquad (33)$$

Expanding

$$\begin{aligned}\|\phi(\mathbf{IDF} \odot \mathbf{z})\|_2^2 &= \sum_k \left(\sum_{i:h(w_i)=k} \mathrm{IDF}_i \mathbf{z}_i\right)^2 \\ \|\overline{\mathbf{IDF}} \odot \phi(\mathbf{z})\|_2^2 &= \sum_k \overline{\mathrm{IDF}}_k^2 \left(\sum_{i:h(w_i)=k} \mathbf{z}_i\right)^2. \end{aligned} \qquad (34)$$

From lemma 1 we have with probability at least $(1 - e^{-1})$, $\mathrm{IDF}(w_i) \geq \overline{\mathrm{IDF}}(w_i) \geq \mathrm{IDF}(w_i) - \epsilon \|\mathbf{f}\|$, provided $m \geq \lceil \exp(1)/\epsilon \rceil$. Here $\overline{\mathrm{IDF}}(w_i) = \overline{\mathrm{IDF}}_k$ for any $w_i$ such that $h(w_i) = k$. Fix $i$ and the corresponding $k$. We have $\mathrm{IDF}_i \geq \overline{\mathrm{IDF}}_k \geq \mathrm{IDF}_i - \epsilon \|\mathbf{f}\|$. Substituting $\mathrm{IDF}_i \geq \overline{\mathrm{IDF}}_k$ leads to $\left(\|\phi(\mathbf{IDF} \odot \mathbf{z})\|_2^2 - \|\bar{\mathbf{x}}\|_2^2\right) > 0$. Thus $\|\mathbf{x}\|_2^2 - \|\bar{\mathbf{x}}\|_2^2 > \epsilon \|\mathbf{x}\|_2^2$ with probability at least $(1 - e^{-1})(1 - 3\delta)$. Using $\mathrm{IDF}_i \leq \overline{\mathrm{IDF}}_k + \epsilon \|\mathbf{f}\|$ lead to

$$\left(\|\phi(\mathbf{IDF} \odot \mathbf{z})\|_2^2 - \|\bar{\mathbf{x}}\|_2^2\right) \leq \sum_k \left(2\epsilon \|\mathbf{f}\| \overline{\mathrm{IDF}}_k + \epsilon^2 \|\mathbf{f}\|^2\right) \left(\sum_{i:h(w_i)=k} \mathbf{z}_i\right)^2. \qquad (35)$$

Since $\|\mathbf{f}\|$ is the sum of all the document frequencies and $\mathrm{IDF}_i \leq 1$, we can assume that $\overline{\mathrm{IDF}}_k \leq \|\mathbf{f}\|$. Using also that $\epsilon^2 \leq \epsilon$ for any $\epsilon \in (0, 1)$, we obtain,

$$\left(\|\phi(\mathbf{IDF} \odot \mathbf{z})\|_2^2 - \|\bar{\mathbf{x}}\|_2^2\right) \leq 3\epsilon \|\mathbf{f}\|^2 \sum_k \left(\sum_{i:h(w_i)=k} \mathbf{z}_i\right)^2 = 3\epsilon \|\mathbf{f}\|^2 \|\phi(\mathbf{z})\|_2^2. \qquad (36)$$

Using (32) and $\epsilon^2 \leq \epsilon$ again we have

$$\left(\|\phi(\mathbf{IDF} \odot \mathbf{z})\|_2^2 - \|\bar{\mathbf{x}}\|_2^2\right) \leq 6\epsilon \|\mathbf{f}\|^2 \|\mathbf{x}\|_2^2. \qquad (37)$$

Combining with (32) and since $\|\mathbf{f}\| > 1$ (at least one word in one document was observed), we have that with probability at least $(1 - e^{-1})(1 - 3\delta)$, $\|\mathbf{x}\|_2^2 - \|\bar{\mathbf{x}}\|_2^2 < 7\epsilon \|\mathbf{f}\|^2 \|\mathbf{x}\|_2^2$. To obtain the result, set $\bar{\epsilon} = 7\epsilon \|\mathbf{f}\|^2$ and substitute in Theorem 1. Note that $\epsilon$ can be arbitrarily small and so $\bar{\epsilon}$. □

*Proof.* (Proposition 2) Clearly,

$$\Pr(\text{Err}) \leq \Pr(\overline{\text{Err}}) + \Pr_{\mathbf{x},\boldsymbol{\phi}}\left(f_{\mathcal{X}}(\mathbf{x}) \neq f_{\boldsymbol{\phi}}(\mathbf{x})\right), \tag{38}$$

where $\Pr_{\mathbf{x},\boldsymbol{\phi}}\left(f_{\mathcal{X}}(\mathbf{x}) \neq f_{\boldsymbol{\phi}}(\mathbf{x})\right)$ is the probability that the embedding changes a decision made in the original space. This probability is taken both with respect to the randomness of $\boldsymbol{\phi}(\cdot)$ and $\mathbf{x}$. Let $I(\xi)$ be the indicator function for the event $\xi$, that is $I(\xi) = 1$ if $\xi$ is verified and 0 otherwise.

$$\begin{aligned} \Pr_{\mathbf{x},\boldsymbol{\phi}}\left(f_{\mathcal{X}}(d) \neq f_{\boldsymbol{\phi}}(d)\right) &= E_{\mathbf{x},\boldsymbol{\phi}}\left[I\left(f_{\mathcal{X}}(\mathbf{x}) \neq f_{\boldsymbol{\phi}}(\mathbf{x})\right)\right] \\ &= E_{\mathbf{x}} E_{\boldsymbol{\phi}|\mathbf{x}} I\left(f_{\mathcal{X}}(\mathbf{x}) \neq f_{\boldsymbol{\phi}}(\mathbf{x})\right) \\ &= E_{\mathbf{x}} \Pr_{\boldsymbol{\phi}|\mathbf{x}}(f_{\mathcal{X}}(\mathbf{x}_i) \neq f_{\boldsymbol{\phi}}(\mathbf{x}_i)), \end{aligned} \tag{39}$$

where $E_{\mathbf{x}}$ denotes the mean among the test instances. Now we can average among the instances satisfying $\eta(\mathbf{x}) \geq \eta*$ and the instances violating $\eta(\mathbf{x}) \geq \eta*$,

$$\begin{aligned} E_{\mathbf{x}} \Pr_{\boldsymbol{\phi}|\mathbf{x}}\left(f_{\mathcal{X}}(\mathbf{x}) \neq f_{\boldsymbol{\phi}}(\mathbf{x})\right) =& E_{\mathbf{x},\eta(\mathbf{x}) \geq \eta*} \Pr_{\boldsymbol{\phi}|\mathbf{x}}\left(f_{\mathcal{X}}(\mathbf{x}) \neq f_{\boldsymbol{\phi}}(\mathbf{x})\right) \\ &+ E_{\mathbf{x},\eta(\mathbf{x}) < \eta*} \Pr_{\boldsymbol{\phi}|\mathbf{x}}\left(f_{\mathcal{X}}(\mathbf{x}) \neq f_{\boldsymbol{\phi}}(\mathbf{x})\right). \end{aligned} \tag{40}$$

Clearly,

$$E_{\mathbf{x},\eta(\mathbf{x})<\eta*} \Pr_{\boldsymbol{\phi}|\mathbf{x}}\left(f_{\mathcal{X}}(\mathbf{x}) \neq f_{\boldsymbol{\phi}}(\mathbf{x})\right) \leq E_{\mathbf{x},\eta(\mathbf{x})<\eta*} \mathbf{1} = \delta_{\eta*}. \tag{41}$$

On the other hand, for the instances satisfying $\eta(\mathbf{x}) \geq \eta*$, the model parameters $\bar{\eta}* = \eta*/D^2$, $m \geq 48\bar{\eta}^{-2} \log(3n_t/\bar{\delta})$ and $c = 32\bar{\eta}^{-1} \log(3n_t/\bar{\delta}) \log^2(3mn_t/\bar{\delta})$ are enough to apply Lemma 2. Thus,

$$E_{\mathbf{x},\eta(\mathbf{x})\geq\eta*} \Pr_{\boldsymbol{\phi}|\mathbf{x}}\left(f_{\mathcal{X}}(\mathbf{x}) \neq f_{\boldsymbol{\phi}}(\mathbf{x})\right) \leq E_{\mathbf{x},\eta(\mathbf{x})\geq\eta*}\left[\bar{\delta}\right] = (1 - \delta_{\eta*})\bar{\delta}.$$

If all the data $\phi(D) = \{\mathbf{x}_1, \ldots, \mathbf{x}_{n_t}\}$ is correctly classified with margin $\eta* > 0$ in the original space, $\Pr(\overline{\text{Err}}) = 0$ and $\delta_{\eta*} = 0$. By construction, $m \geq 48\bar{\eta}^{-2} \log(3n_t/\bar{\delta})$. Solving for $\bar{\delta}$ gives,

$$\bar{\delta} \leq 3n_t \exp\left(-\frac{m\eta*^2}{48D^4}\right). \tag{42}$$

□